\documentclass{article}

\usepackage[numbers]{natbib}
\usepackage{algorithm}
\usepackage{algorithmic}
\usepackage{float}  
\usepackage{lipsum}
\usepackage{subfigure}
\usepackage[normalem]{ulem}

\usepackage[table]{ xcolor}
\usepackage{enumitem}
\usepackage{amsmath}

\useunder{\uline}{\ul}{}

\usepackage[preprint]{neurips_2023}

\usepackage{wrapfig}
\usepackage[utf8]{inputenc} 
\usepackage[T1]{fontenc}    
\usepackage{hyperref}       
\usepackage{url}            
\usepackage{booktabs}       
\usepackage{amsfonts}       
\usepackage{nicefrac}       
\usepackage{microtype}      
\usepackage{xcolor}         
\usepackage{graphicx}
\usepackage{amsmath}

\title{GIMM: InfoMin-Max for Automated Graph Contrastive Learning}

\author{%
  Xin Xiong \\
  School of Artificial Intelligence\\
  Nanjing University\\
  \texttt{xiongxin@smail.nju.edu.cn} \\
  \And
  Furao Shen \\
  School of Artificial Intelligence \\
  Nanjing University \\
  \texttt{frshen@nju.edu.cn} \\
  \AND
  Xiangyu Wang \\
  School of Artificial Intelligence\\
  Nanjing University \\
  \texttt{xiangyuwang@smail.nju.edu.cn} \\
  \And
  Jian Zhao \\
  School of Electronic Science and Engineering\\
  Nanjing University \\
  \texttt{jianzhao@nju.edu.cn} \\
}

\begin{document}

\maketitle

\begin{abstract}
  Graph contrastive learning (GCL) shows great potential in unsupervised graph representation learning. Data augmentation plays a vital role in GCL, and its optimal choice heavily depends on the downstream task.
  Many GCL methods with automated data augmentation face the risk of insufficient information as they fail to preserve the essential information necessary for the downstream task.
  To solve this problem, we propose \underline{I}nfo\underline{M}in-\underline{M}ax for automated \underline{G}raph contrastive learning (GIMM), which prevents GCL from encoding redundant information and losing essential information. GIMM consists of two major modules: (1) \textit{automated graph view generator}, which acquires the approximation of InfoMin’s optimal views through adversarial training without requiring task-relevant information; (2) \textit{view comparison}, which learns an excellent encoder by applying InfoMax to view representations. To the best of our knowledge, GIMM is the first method that combines the InfoMin and InfoMax principles in GCL. 
  Besides, GIMM introduces randomness to augmentation, thus stabilizing the model against perturbations. 
  Extensive experiments on unsupervised and semi-supervised learning for node and graph classification demonstrate the superiority of our GIMM over state-of-the-art GCL methods with automated and manual data augmentation.
\end{abstract}

\section{Introduction}\label{intro}
Labeling graphs is a very challenging and laborious task since it generally requires domain knowledge, and graphs usually have numerous nodes with complex relationships.
Thus, unsupervised graph representation learning~\cite{mo2022simple,park2019symmetric,mavromatis2020graph,peng2020graph} has gained significant attention recently, which aims to obtain low-dimensional representations of nodes or graphs without label information. 
These representations can be used for a wide range of downstream tasks, such as node classification~\cite{velickovic2019deep}, graph classification~\cite{sun2019infograph}, and graph clustering~\cite{pan2021multi}.
Graph contrastive learning (GCL)~\cite{you2020graph,you2021graph,zhu2021graph,tong2021directed,zhu2020deep} shows great potential in unsupervised graph representation learning due to its excellent ability of expression, and it generally includes two sequential modules, \textit{view generation} and \textit{view comparison}.
View generation generates two views by data augmentation on the original graph.
View comparison acquires view representations through an encoder and then optimizes the encoder by pulling view representations from the same distribution closer while pushing away view representations from different distributions.
\begin{figure*}
    \centering
     \includegraphics[scale=0.49]{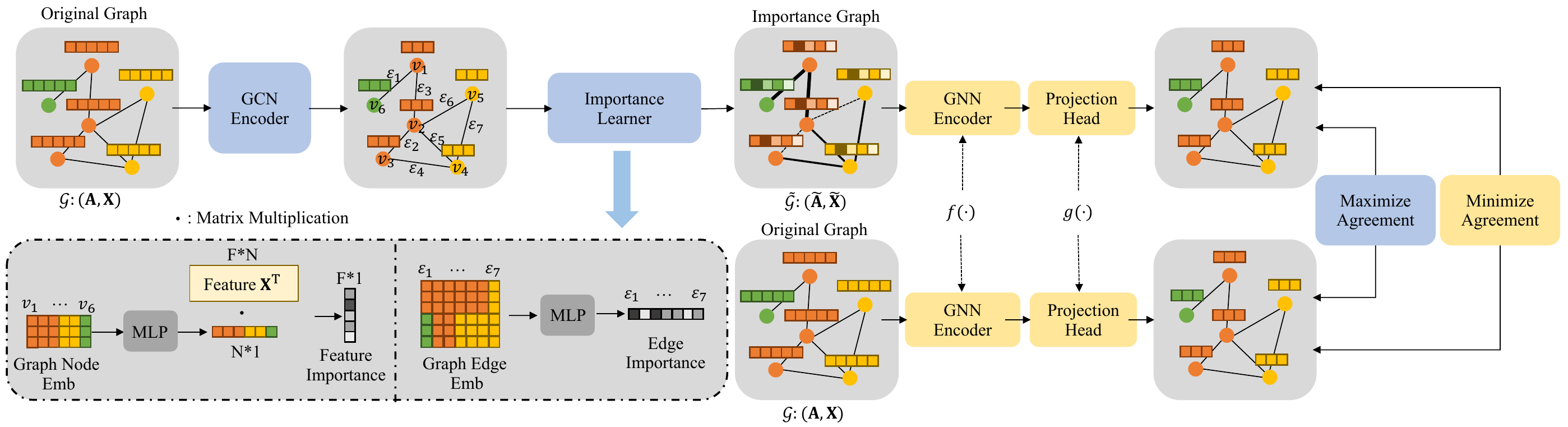}
     \caption{Overview of the automated graph view generator. 
     In the importance graph, darker features and thicker edges are more significant, whereas dashed edges are less vital.
     The original graph is fed into a GCN encoder and an importance learner to generate an importance graph.
     After acquiring representations of the importance and original graph using a shared GNN encoder and projection head, max-min optimization is performed.
     Through max-min optimization, the importance graph can emphasize the minimal noteworthy information.
     The views defined by the importance graph approximate the optimal views in InfoMin.}
     \label{fig:view_gen}
  \end{figure*}

Nevertheless, GCL heavily depends on data augmentation, and an inappropriate augmentation will lead to severe performance loss. 
Graph data augmentation (GDA) necessitates the consideration of both the complex graph topology and feature information. Various GDA techniques have been proposed, which can be categorized into three categories based on their augmentation modality~\cite{ding2022data}: structure-oriented~\cite{DBLP:journals/corr/abs-2001-07524,velickovic2019deep, zhu2021graph,page1999pagerank,kondor2002diffusion}, feature-oriented~\cite{feng2019graph,yang2021graph,you2020graph}, and label-oriented~\cite{you2020does,park2022graph}.
Ensuring the selection of an appropriate GDA technique is critical in GCL~\cite{you2020graph}, prompting many GCL methods to rely on trial-and-error or empirical approaches when choosing data augmentations.
GraphCL~\cite{you2020graph} introduces four GDA techniques, and demonstrates that good GDA relies on specific characteristics of various graph data through pairwise combinations of these data augmentations.
MERIT~\cite{jin2021multi} achieves data augmentation by superimposing the methods from its data augmentation pool.
GCA~\cite{zhu2021graph} conducts data augmentation by employing three centralities and selecting the optimal one according to its performance on the downstream task.
On the one hand, the methods above require a prudent design of the data augmentation pool. On the other hand, selecting the best data augmentation according to performance on the downstream task of various datasets is entailed, which is both time-consuming and computationally expensive. Thus, it is significantly advantageous to automate GDA.
\begin{figure}
   \setlength{\belowcaptionskip}{-0.3cm}
    \centering
     \includegraphics[scale=0.54]{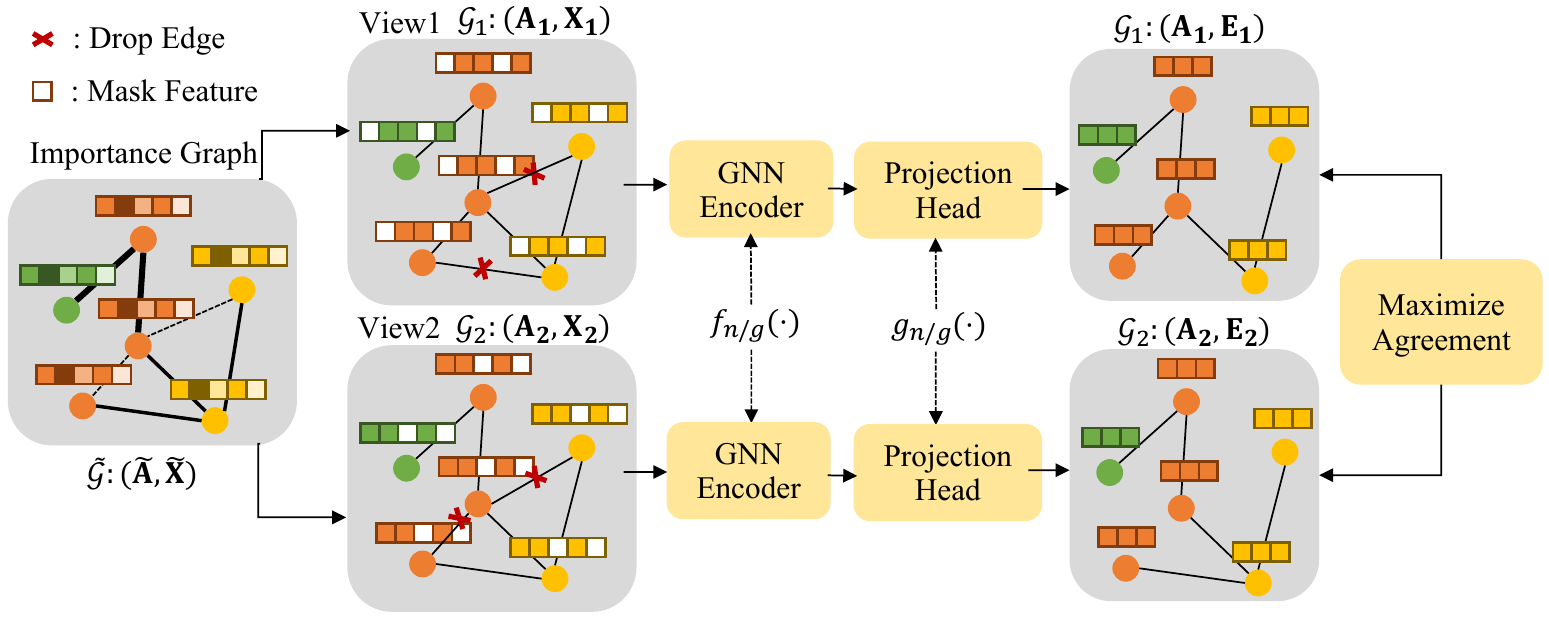}
     \caption{Overview of the view comparison module. Initially, two views are generated by masking features and dropping edges according to the importance graph. Representations of the views are calculated through a shared GNN encoder and projection head. We adopt InfoNCE to maximize the mutual information between the two representations to guide the model to learn the basic topology and feature information in the graph. If the downstream task is graph classification, the node representations need to pass through a readout function to get graph representations.}
     \label{fig:view_comp}
  \end{figure}

The mutual information maximization principle (InfoMax)~\cite{linsker1988self} is widely used in GCL~\cite{velickovic2019deep,zhu2020deep,thakoor2021bootstrapped}, which refers to maximizing the agreement (mutual information, MI) between view representations. 
Through InfoMax, the contrastive model can identify the pairs augmented from the same node or graph, thereby learning the basic topology and feature information on the graph.
However, InfoMax may risk the model learning redundant information irrelevant to the downstream task but beneficial for identifying the pairs.
Encoding redundant information results in brittle representations and may severely degrade the encoder's performance on the downstream task ~\cite{tschannen2019mutual}.
For example, when training an optical character recognition model, the color information is redundant and harmful to recognition, and should be removed beforehand.
Thus, before implementing InfoMax, redundant information shared between views should be stripped out.
Now the new question is \textit{what kind of views InfoMax needs.}
The InfoMin principle~\cite{tian2020makes} in the visual domain indicates that a good set of views should share the minimal information necessary to perform well at the downstream task, and we name such information \textit{minimal necessary information}.
It points out that information between views should contain the necessary information needed for the downstream task while excluding nuisance information ~\cite{tian2020makes}.
Indeed, GCL methods with manual data augmentation essentially search in the data augmentation pool for augmentations that yield views that best satisfy the InfoMin principle.
However, many GCL methods with automated data augmentation fail to follow InfoMin fully.
AutoGCL~\cite{yin2022autogcl}, AD-GCL~\cite{suresh2021adversarial}, and JOAO~\cite{you2021graph} adhere to the min-max optimization framework, seeking views with minimal similarity or seeking views that are the most challenging.
Unfortunately, the views derived by minimizing the similarity between views or by minimizing the agreement between view representations are far from optimal. 
These approaches reduce nuisance information but do not emphasize the retention of the necessary information, potentially resulting in the excessive removal of information and risking the model suffering from insufficient information. 
Therefore, applying InfoMin to acquire optimal views becomes the key: removing redundant information while maintaining necessary information related to the downstream task.
However, it is impossible to measure task-relevant information under an unsupervised setting. Thus, it is hard to define minimal necessary information in InfoMin.
Nevertheless, we find minimal information that is noteworthy for different downstream tasks can approximate minimal necessary information, which avoids requiring task-relevant information.
We name such information \textit{minimal noteworthy information} (MNI). 

In this paper, we propose a novel method called \underline{I}nfo\underline{M}in-\underline{M}ax for automated \underline{G}raph contrastive learning (GIMM).
Specifically, GIMM uses max-min optimization to learn an importance graph. The set of views defined by the importance graph shares MNI, making these views approximations of InfoMin's optimal views.
In addition, GIMM introduces randomness to views, thereby stabilizing the model against perturbations.
Finally, an encoder is optimized by applying InfoMax to view representations.
GIMM achieves outstanding performances without using corrupted views, showing that negative views are unnecessary for GCL.

\textbf{Our contributions.}
(i) A novel unsupervised GCL method with automated data augmentation, GIMM, is proposed. To the best of our knowledge, GIMM is the first method that combines InfoMin and InfoMax principles in GCL.
(ii) A better approximation of the InfoMin principle in unsupervised graph representation learning is achieved.
We employ an adversarial training strategy to generate views that share minimal noteworthy information, which avoids using task-relevant information. Applying InfoMax to these views is risk-free, as they reduce nuisance information and emphasize noteworthy information to ensure sufficient information. In addition, a theoretical motivation is provided.
(iii) Extensive experiments on node and graph classification demonstrate the effectiveness of our approximately optimal views on different tasks and the superiority of GIMM over state-of-the-art (SOTA) GCL methods with automated and manual data augmentation.
\section{Related work}
Matrix-factorization-based~\cite{ahmed2013distributed,cao2015grarep,ou2016asymmetric} and random-walk-based methods~\cite{perozzi2014deepwalk,grover2016node2vec} are classical approaches for unsupervised graph representation learning. However, these methods may emphasize topology excessively while neglecting feature information.
Deep unsupervised graph representation learning~\cite{zhou2022multiview,oyallon2020interferometric} has gradually evolved in recent years, with GCL emerging as a promising approach.
InfoMax~\cite{linsker1988self} is one of the most commonly used principles in GCL.
DGI~\cite{velickovic2019deep} follows DIM~\cite{hjelm2018learning}, applying InfoMax to graph data for the first time.
It uses node shuffling to corrupt the original graph to generate negative pairs, and the positive and negative pairs are distinguished through a discriminator optimized by InfoMax. 
GRACE~\cite{zhu2020deep} generates two views through the corruption of removing edges and masking node features and then learns representations by applying InfoMax.
BGRL~\cite{thakoor2021bootstrapped} generates two views through stochastic node feature masking and edge masking, and learns by maximizing the cosine similarity of the view representations at different stages, essentially applying InfoMax.
All the above works involve manually selecting data augmentation or corruption method. However, the corruption method is not necessary for contrastive learning~\cite{thakoor2021bootstrapped}. Therefore, our work focuses on GCL with automated data augmentation.

JOAO~\cite{you2021graph} searches for the best combination of data augmentations from a fixed pool, with the combination coefficients determined by min-max optimization. 
However, designing a proper data augmentation pool still involves human knowledge, and only optimizing the combination coefficient limits the flexibility of data augmentation.
AD-GCL~\cite{suresh2021adversarial} augments graphs via adversarial training following the information bottleneck (IB) principle ~\cite{tishby2000information,goldfeld2020information,alemi2016deep}, hence reducing redundant information shared between views.
In fact, IB and InfoMin are closely related.
IB states that the encoder should minimize the information in the original data while maximizing information relevant to downstream tasks.
Still, AD-GCL only augments on edges but ignores critical feature information.
In InfoGCL~\cite{cheng2022information}, data augmentation is decided mathematically by minimizing mutual information between views and maximizing mutual information between views and tasks, which essentially follows IB.
However, a manually designed data augmentation pool is entailed in InfoGCL too, and the best augmentation is acquired by exhaustive search, which is time-consuming and computationally expensive.
AutoGCL~\cite{yin2022autogcl} uses two learnable view generators, each of which learns a probability distribution over the nodes of the input graph.
It minimizes the similarity between two views and maximizes the agreement between the representations of the two views.
Nevertheless, it merely executes node-level view learning without considering edges.

\section{Methodology}
GIMM consists of two sequential modules, the \textit{automated graph view generator} and the \textit{view comparison module}.
The automated graph view generator acquires an importance graph used to generate two views. These two views are fed into the view comparison module.
Figure \ref{fig:view_gen} and \ref{fig:view_comp} illustrate the structures of the two modules.

\textbf{Notations.}
A graph $\mathcal{G}$ consists of $\mathcal{V}=\{v_1,...,v_N \}$, a set of nodes and $\mathcal{E} = \{\varepsilon_1,...,\varepsilon_M\}$, a set of edges. 
The feature matrix, adjacency matrix and degree matrix of $\mathcal{G}$ are denoted by $\textbf{X} = \{\textbf{x}_1,...,\textbf{x}_N\} \in \mathbb{R}^{N\times F}$, $\textbf{A}\in \{0,1\}^{N\times N}:\textbf{A}_{ij}=\mathbb{I}((v_i,v_j)\in\mathcal{E})$ and $\textbf{D}$: $\textbf{D}_{ii} = \sum_j \textbf{A}_{ij}$, where $F$ is the feature dimension and $\textbf{x}_i \in \mathbb{R}^F$ is
the feature of node $v_i$.
We use $\mathcal{G}: (\textbf{A},\textbf{X})$ to represent the graph.
GIMM achieves the approximation of the InfoMin principle without task-relevant information. To verify the effectiveness of such an approximation strategy for different downstream tasks, we discuss two downstream tasks, node classification and graph classification.
For node classification, whose input is a single graph $\mathcal{G}$,
we aim to learn an encoder $f_n: \mathcal{G}\rightarrow\mathbb{R}^{N\times d}$ to obtain the low-dimensional node embeddings $f_n(\mathcal{G}),d\ll F$.
For graph classification, whose input is a set of graphs $\{\mathcal{G}_i\}_{i = 1}^Q$, we aim to learn an encoder $f_g: \mathcal{G}_i \rightarrow \mathbb{R}^{d}$ to obtain the low-dimensional graph embeddings $f_g(\mathcal{G}_i), i = 1,...,Q$.

\textbf{InfoMin and InfoMax principle.}
$X$, $Y$ denote the input and downstream task information. $V_1$, $V_2$ denote the views, and $f$ denotes the encoder.
InfoMin principle, which seeks the optimal views, states that
$(V_1^*,V_2^*)=\operatorname{argmin}_{V_1,V_2} I(V_1;V_2),$ subject to
$I(V_1;Y)=I(V_2;Y)=I(X;Y),$ where $I(V_1;V_2)$ is the mutual information between $V_1$ and $V_2$.
InfoMax principle learns the encoder via maximizing the mutual information between view representations, i.e., $\max_f I(f(V_1);f(V_2))$.

\textbf{Min-max or max-min?}
\bibinfo{author}{Tian} et al.~\cite{tian2020makes} leverage a min-max training strategy to complete unsupervised view learning of InfoMin. 
Given image $X$, the transformed image $\hat{X}=g(X)$ and two encoders $f_1,f_2$, the objective is:
\begin{equation}
    \min_g \max_{f_1,f_2} I(f_1(g(X)_1);f_2(g(X)_{2:3})). 
\end{equation}
$\{g(X)_1,g(X)_{2:3}\}$ represent the split channels of $g(X)$ and thus serve as the two views $V_1,V_2$.
The authors mention that this strategy heavily breaks constraint $I(V_1;Y)=I(V_2;Y)=I(X;Y)$, and $I(V_1;V_2)$ is overly reduced.
The objective poses the risk of insufficient information, as Section \ref{intro} mentions.
Other methods, such as JOAO, AD-GCL, and AutoGCL, follow this min-max training strategy, reducing redundant information but potentially resulting in insufficient information.
Therefore, we propose a max-min training strategy to reduce nuisance information and preserve noteworthy information simultaneously.

\subsection{Automated graph view generator}
Applying the InfoMin principle under an unsupervised setting is challenging since it is impossible to measure task-relevant information, as stated in Section \ref{intro}.
Nevertheless, we propose seeking views that share minimal information noteworthy for different downstream tasks, which is the approximation of the InfoMin principle and without requiring task-relevant information. 
We name such information \textit{minimal noteworthy information} (MNI). 
The information contained in a graph comprises both its topology and features. Therefore, MNI is a subset of crucial edges and features.

An importance graph refers to a graph with edge and feature importance.
The importance graph of $\mathcal{G}: (\textbf{A},\textbf{X})$ is denoted by $\tilde{\mathcal{G}} = T(\mathcal{G}) = (\tilde{\textbf{A}}, \tilde{\textbf{X}})$. 
MNI is those edges and features with high importance.
Our optimization strategy for the automated graph view generator is
\begin{equation}
\max_{T}\min_{f,g}I(g(f(\mathcal{G}));g(f(T(\mathcal{G})))) + \min_{T} |T(\mathcal{G})|,\ \operatorname{s.t.} I(g(f(\mathcal{G}));g(f(T(\mathcal{G}))))\geq\zeta.\label{equa:minmax}
\end{equation}
where $f$ is a graph encoder for node embedding, and $g$ is a projection head to increase the ability of expression.
It should be mentioned that $f,g$ have nothing to do with the encoder of the view comparison module.
$I(g(f(\mathcal{G}));g(f(T(\mathcal{G}))))\geq\zeta$ is introduced to prevent the degeneration of $g(f(\cdot))$ into the corner case, i.e., $I(g(f(\mathcal{G}));g(f(T(\mathcal{G}))))=0$.
However, finding such a corner case is challenging for the optimizer, and thus we do not incorporate this constraint in our implementation. Experiments indicate that such a corner case does not occur.
The regularization term $|T(\mathcal{G})|$ is the normalized sum of edge importance and feature importance.
The intuition of our strategy is in the most challenging case, i.e., using a very aggressive information encoder $g(f(\cdot))$, the emphasized information $T(\mathcal{G})$ is more critical. Next, we give the theoretical motivation.

\textbf{The critical information in the graph is possibly highly relevant to various downstream tasks.}
For example, the critical edges connecting different clusters within a graph play a crucial role in tasks like graph partition~\cite{karypis1998fast,nazi2019gap}, graph classification~\cite{kipf2016semi,han2022g}, and link prediction~\cite{zhang2018link,yang2021inductive}. These edges serve as essential connections that provide valuable insights into the underlying relationships and dependencies within the graph.
MNI is recognized as critical information in a graph and is generally task-relevant.
Experiments in Section \ref{exp} reveal the effectiveness of MNI in both node and graph classification.
\textbf{The objective of our automated graph view generator is approximately equivalent to decreasing $I(\mathcal{G};\tilde{\mathcal{G}}|Y)$ and increasing $I(\tilde{\mathcal{G}};Y)$.}
We have
\begin{equation}
    I(g(f(\mathcal{G}));g(f(\tilde{\mathcal{G}}))|Y)
        =I(g(f(\mathcal{G}));g(f(\tilde{\mathcal{G}})))-I(g(f(\tilde{\mathcal{G}}));Y)+I(g(f(\tilde{\mathcal{G}}));Y|g(f(\mathcal{G}))).
\end{equation}
$0 \overset{(a)}{\leq} I(g(f(\tilde{\mathcal{G}}));Y|g(f(\mathcal{G})))\overset{(b)}{\leq} I(g(f(\mathcal{G}));Y|g(f(\mathcal{G})))=0$.
$(a)$ is because mutual information is non-negative, while $(b)$ is due to the data processing inequality~\cite{cover1999elements} and $\tilde{\mathcal{G}}$ is a function of $\mathcal{G}$. Thus,
\begin{equation}
    I(g(f(\mathcal{G}));g(f(\tilde{\mathcal{G}}))|Y)+I(g(f(\tilde{\mathcal{G}}));Y)
    =I(g(f(\mathcal{G}));g(f(\tilde{\mathcal{G}}))).
\label{eq:Theor}
\end{equation}
The right side of Eqn. \ref{eq:Theor} is exactly our max-min term in Eqn. \ref{equa:minmax}. 
According to Eqn. \ref{equa:minmax}, the optimal solution for $\max_{T}I(g(f(\mathcal{G}));g(f(\mathcal{\tilde{G}}))),\mathcal{\tilde{G}}=T(\mathcal{G})$ is to set $\mathcal{G} = \mathcal{\tilde{G}}$.
However, as $\min_T|T(\mathcal{G})|$ serves as a regularization, the number of emphasized edges and features is limited.
Thus, minimal noteworthy information consists of two parts: minimal information for $\min_{T} |T(\mathcal{G})|$ and noteworthy information for $\max_{T}I(g(f(\mathcal{G}));g(f(T(\mathcal{G}))))$.
Emphasizing the most critical edges and features can maximize the mutual information between the importance graph $\mathcal{\tilde{G}}$ and the original graph $\mathcal{G}$, and those edges and features are possibly highly task-relevant.
As the optimization advances, the information within $\mathcal{\tilde{G}}$ becomes increasingly ``compact'' and ``critical.''
Consequently, there is an increased proportion of task-relevant information $I(g(f(\tilde{\mathcal{G}}));Y)$ and a decreased proportion of task-irrelavant information $I(g(f(\mathcal{G}));g(f(\tilde{\mathcal{G}}))|Y)$.
Therefore, the increased $I(\tilde{\mathcal{G}};Y)$ and decreased $I(\mathcal{G};\tilde{\mathcal{G}}|Y)$ are achieved under the limited information of $\tilde{\mathcal{G}}$.
The objective of $\min_{f,g}I(g(f(\mathcal{G}));g(f(\mathcal{\tilde{G}}))|Y)$ is to obtain a challenging information encoder $g(f(\cdot))$.
It is expected that the MNI obtained under a challenging information encoder $g(f(\cdot))$ will exhibit higher robustness, and the experiments comparing GIMM and GIMM-ViewM have demonstrated the effectiveness of minimization (see Section \ref{exp} for more details).
Ultimately, the views generated using the importance graph $\tilde{\mathcal{G}}$ approximate the goal of preserving task-relevant information while discarding task-irrelevant information.

\subsubsection{Importance graph}
This section describes the construction of the importance graph.
First, node representations are computed using a graph convolutional layer~\cite{welling2016semi}:
$ \textbf{E} = \sigma(\textbf{A}\textbf{X}\textbf{W}),$
where $\sigma$ is the activation function, $\textbf{W}$ is the parameter matrix.
$\textbf{A}$ is the adjacency matrix without normalization instead of the Laplacian matrix $\textbf{D}^{-1/2}\textbf{A}\textbf{D}^{-1/2}$, as the node degrees are also essential for importance computation.
We need to get the node importance first to get the feature importance.
The node importance $\textbf{P}_n \in \mathbb{R}^{N\times 1}$ is given by $\textbf{P}_n = \operatorname{Gumbel}(h_\phi(\textbf{E})),$
where $h_\phi(\cdot)$ is a simple MLP, and $\operatorname{Gumbel}(t) = \operatorname{Sigmoid}((\log \eta-\log(1-\eta)+t)/{\tau}),\ \eta \sim\operatorname{Uniform}(0,1)$ is the Gumbel-Max reparametrization function.
$\tau$ is the temperature parameter, and the closer $\tau$ is to 0, the closer $\textbf{P}_n$ is to binarization. The Gumbel-Max reparametrization trick~\cite{maddison2016concrete} ensures the process can be backpropagated and makes $\textbf{P}_n$ probabilistically meaningful.
Features with larger values in important nodes are generally more critical. Thus, feature importance $\textbf{P}_f \in \mathbb{R}^{F\times 1}$ can be calculated through node importance, i.e., 
$
    \textbf{P}_f = \textbf{X}^T \textbf{P}_n,
$
where $\textbf{X}^T\in \mathbb{R}^{F\times N}$ is the transpose of feature matrix $\textbf{X}$. 
If features are real numbers between 0 and 1, we perform the Gumbel-Max reparametrization trick on $\textbf{P}_f$ rather than on $\textbf{P}_n$ to avoid $\textbf{P}_f$ being too small.

Inspired by AD-GCL~\cite{suresh2021adversarial}, 
the importance of edge $\varepsilon_k =(v_i,v_j)\in \mathcal{E}$ can be expressed by
$\textbf{p}_{e,k} = \operatorname{Gumbel}(h_\psi([\textbf{E}[v_i];\textbf{E}[v_j]])),$
where $[\cdot;\cdot]$ is the concatenation operation, $h_\psi(\cdot)$ is a simple MLP, and $\textbf{E}[v_i]$ is the embedding of node $v_i$.
The edge importance $\textbf{P}_e= [\textbf{p}_{e,1},...,\textbf{p}_{e,M}]^T\in \mathbb{R}^{M\times 1}$.

Broadcast $\textbf{P}_f\in \mathbb{R}^{F\times 1}$ to $\textbf{P}'_f\in \mathbb{R}^{N\times F}$.
The importance graph is $\tilde{\mathcal{G}} = (\tilde{\textbf{A}}, \tilde{\textbf{X}})$. $\tilde{\textbf{X}}$ is given by
$
    \tilde{\textbf{X}} = \textbf{X}\odot \textbf{P}'_f,
$
where $\odot$ is the Hadamard product. We derive $\tilde{\textbf{A}}$ by replacing the edge values in the adjacency matrix $\textbf{A}$ with the edge values of $\textbf{P}_e$.

InfoNCE~\cite{oord2018representation}, a lower bound of mutual information, is applied to estimate mutual information.
The representation of node $v_i$ in $\mathcal{G}$ is defined by $\textbf{z}_{i,1}=g_\xi(f_\theta(\mathcal{G}(v_i)))$ and the representation of it in $\tilde{\mathcal{G}}$ by $\textbf{z}_{i,2}=g_\xi(f_\theta(\tilde{\mathcal{G}}(v_i)))$. 
For simplicity, $f_\theta$ is a 1 or 2- layer graph encoder, and $g_\xi$ is a 1 or 2- layer MLP.
The mutual information between $\textbf{z}_{i,1}$ and $\textbf{z}_{i,2}$ can be estimated by $\hat{I}(\textbf{z}_{i,1};\textbf{z}_{i,2})$:
\begin{equation}
\hat{I}(\textbf{z}_{i,1};\textbf{z}_{i,2})=\frac{1}{2}(\hat{I}_0(\textbf{z}_{i,1};\textbf{z}_{i,2})+\hat{I}_0(\textbf{z}_{i,2};\textbf{z}_{i,1})),
        \hat{I}_0(\textbf{z}_{i,1};\textbf{z}_{i,2})=\log \frac{\exp(s(\textbf{z}_{i,1},\textbf{z}_{i,2})/\epsilon)}{\sum_{j=1,j\neq i}^N \exp(s(\textbf{z}_{i,1},\textbf{z}_{j,2})/\epsilon)},
    \label{equa:infonce}   
\end{equation}

where $s(\cdot,\cdot)$ is the cosine similarity, $\epsilon$ is the temperature parameter, and the symmetrical design considers the same status of $\mathcal{G}$ and $\tilde{\mathcal{G}}$.
The mutual information between the representation of $\mathcal{G}$ and $\tilde{\mathcal{G}}=T(\mathcal{G})$ is estimated by
\begin{equation}
    \setlength\abovedisplayskip{2pt}
    \setlength\belowdisplayskip{-1pt}
    I(g_\xi(f_\theta(\mathcal{G}));g_\xi(f_\theta(T(\mathcal{G})))) \to \hat{I}(g_\xi(f_\theta(\mathcal{G}));g_\xi(f_\theta(T(\mathcal{G}))))
    =\frac{1}{N}\sum_{i=1}^N \hat{I}(\textbf{z}_{i,1};\textbf{z}_{i,2}).
\end{equation}
The regularization term $|T(\mathcal{G})|$ is defined by
$ |T(\mathcal{G})|=\lambda\left(\sum_{i=1}^F\textbf{P}_f^{(i)}/F+ \sum_{i=1}^M\textbf{P}_e^{(i)}/M \right),$

where $\lambda$ is the regularization weight, and $\textbf{P}_f^{(i)}$ is the $i$-th dimension of $\textbf{P}_f$. Finally, our optimization strategy for the automated view generator is
\begin{equation}
    \setlength\abovedisplayskip{1pt}
    \setlength\belowdisplayskip{-1pt}
    \begin{aligned}
    &\max_{\textbf{W},\phi,\psi}\min_{\theta,\xi}\hat{I}(g_\xi(f_\theta(\mathcal{G}));g_\xi(f_\theta(T_{\textbf{W},\phi,\psi}(\mathcal{G})))) +
    \min_{\textbf{W},\phi,\psi} \lambda(\sum_{i=1}^F\textbf{P}_f^{(i)}/ F+\sum_{i=1}^M\textbf{P}_e^{(i)}/ M), \\
    &\operatorname{s.t.}\ \hat{I}(g_\xi(f_\theta(\mathcal{G}));g_\xi(f_\theta(T(\mathcal{G}))))\geq\zeta.
    \end{aligned}
    \label{equa:minmax2}
\end{equation}

\subsubsection{View generation}
The feature importance $\textbf{P}_f$ and edge importance $\textbf{P}_e$ are calculated in the previous section.
MNI is these features and edges with high importance, which are kept in the original graph to generate views. 
These views share MNI and are the approximation of InfoMin's optimal views.
Randomness is introduced in view to stabilizing GIMM against perturbations.
Two views are required, $\mathcal{G}_1:(\textbf{A}_1,\textbf{X}_1)$ and $\mathcal{G}_2:(\textbf{A}_2,\textbf{X}_2)$. Take the generation of $\mathcal{G}_1$ as an example, and the generation of $\mathcal{G}_2$ is similar. 
View modifies both the topology and features of the input graph.
The edge set $\mathcal{E}_1$ of $\mathcal{G}_1$ is a subset of $\mathcal{E}$.
We use a random variable $p_{\varepsilon_k} \sim \operatorname{Bernoulli}(1-p_{d,{\varepsilon_k}})$ to select the edge ${\varepsilon_k}$, i.e., if $p_{\varepsilon_k}=1$, then ${\varepsilon_k}\in \mathcal{E}_1$, else ${\varepsilon_k}\notin \mathcal{E}_1$. 
Therefore, $\textbf{A}_1$ can be derived via $\mathcal{E}_1$.
Inspired by GCA~\cite{zhu2021graph}, $\textbf{P}_{d,\varepsilon}=[p_{d,\varepsilon_1},...,p_{d,\varepsilon_M}]$ is given by
$
    \textbf{P}_{d,\varepsilon} = \min(((\textbf{P}_e^{\max}-\textbf{P}_e)/(\textbf{P}_e^{\max}-\textbf{P}_e^{\operatorname{avg}}))\cdot p_{s1},p_t),
$
where $\min(a,b)$ indicates to select the smaller one of $a$ and $b$, $p_{s1}$ and $p_t$ are hyperparameters between 0 and 1, $p_{s1}$ is used to adjust the overall drop rate of edges, and $p_t$ is used to truncate the drop rate such that every edge has a chance of being included.
$\textbf{P}_e^{\max}, \textbf{P}_e^{\operatorname{avg}}$ are the maximum and average value of $\textbf{P}_e$ respectively. Thus, edges with higher importance have a greater chance of being included in $\mathcal{E}_1$.
\begin{table}[h]
    \centering
    \caption{Node classification datasets.}
    \scalebox{0.76}{
    \begin{tabular}{@{}lrrrr@{}}
    \toprule[1pt]
    Dataset          & \#Nodes    & \#Edges     & \#Features   & \#Classes \\ \midrule
    Wiki-CS          & 11,701 & 216,123 & 300   & 10  \\
    Amazon-Computers & 13,752 & 245,861 & 767   & 10  \\
    Amazon-Photo     & 7,650  & 119,081 & 745   & 8   \\
    Coauthor-CS     & 18,333 & 81,894  & 6,805 & 15  \\
    Coauthor-Physics & 34,493 & 247,962 & 8,415 & 5   \\ \bottomrule[1pt] 
\end{tabular}}
\label{tab:node_clf_data}
\end{table}

\begin{table}[h]
    \centering
    \caption{Graph classification datasets.}
    \scalebox{0.76}{
        \begin{tabular}{@{}lrrrr@{}}
            \toprule[1pt]
            Dataset         & Avg. \#Graphs & Avg. \#Nodes & Avg. \#Edges & \#Classes \\ \midrule
            MUTUG           & 188           & 17.93        & 19.79        & 2         \\
            PROTEINS        & 1,113         & 39.06        & 72.82        & 2         \\
            DD              & 1,178         & 284.32       & 715.66       & 2         \\
            NCI1            & 4,110         & 29.87        & 32.30        & 2         \\
            COLLAB          & 5,000         & 74.49        & 2457.78      & 3         \\
            GITHUB          & 12,725        & 113.79       & 234.64       & 2         \\
            IMDB-BINARY     & 1,000         & 19.77        & 96.53        & 2         \\
            REDDIT-BINARY   & 2,000         & 429.63       & 497.75       & 2         \\
            REDDIT-MULTI-5K & 4,999         & 508.52       & 594.87       & 5         \\ \bottomrule[1pt]
            \end{tabular}}
    \label{tab:graph_clf_data}
    \end{table}

\begin{table}[h]
    \centering
    \setlength{\belowcaptionskip}{0.35cm}
    \caption{   
    Unsupervised learning performance on (TOP) node classification and (BOTTOM) graph classification on the benchmark TUDataset~\cite{morris2020tudataset} (average accuracy $\pm$ std. over 5 runs). Except for AD-GCL,
    JOAO-v2, AutoGCL (on node classification) and GCA (on graph classification), the experimental results of baselines are from published papers. `-' indicates that results are unavailable in papers. OOM indicates Out-Of-Memory on a 32GB GPU.
    The highest and second performances are in \textbf{bold} and {\ul underlined} respectively.}
    \scalebox{0.62}{
        \begin{tabular}{@{}llrrrrr@{}}
            \toprule[1pt]
            Type                      & Model   & Wiki-CS~\cite{mernyei2020wiki} & Amaz-Comp~\cite{shchur2018pitfalls} & Amaz-Photo~\cite{shchur2018pitfalls} & Coauthor-CS~\cite{shchur2018pitfalls} & Coauthor-Phy~\cite{shchur2018pitfalls} \\ \midrule
            w/o GDA   & DGI~\cite{velickovic2019deep}     & 75.35 $\pm$ 0.14                                     & 83.95 $\pm$ 0.47                                                 & 91.61 $\pm$ 0.22                                             & 92.15 $\pm$ 0.63                                            & 94.51 $\pm$ 0.52                                                 \\
                                      & GMI~\cite{peng2020graph}     & 74.85 $\pm$ 0.08                                     & 82.21 $\pm$ 0.31                                                 & 90.68 $\pm$ 0.17                                             & OOM                                                         & OOM                                                              \\ \midrule
            w/ Manual GDA  & MVGRL~\cite{hassani2020contrastive}   & 77.52 $\pm$ 0.08                                     & 87.52 $\pm$ 0.11                                                 & 91.74 $\pm$ 0.07                                             & 92.11 $\pm$ 0.12                                            & 95.33 $\pm$ 0.03                                                 \\
                                      & GCA~\cite{zhu2021graph}     & {\ul 78.35 $\pm$ 0.05}              &  {\ul 87.85 $\pm$ 0.31}                          &  {\ul 92.53 $\pm$ 0.16}                      &  {\ul 93.10 $\pm$ 0.01}                     &  {\ul  95.73 $\pm$ 0.03}                          \\ \midrule
            w/ Automated GDA & AD-GCL~\cite{suresh2021adversarial}  & 73.46 $\pm$ 0.36                                     & 81.32 $\pm$ 0.93                                                 & 88.75 $\pm$ 0.92                                             & 92.16 $\pm$ 0.36                                            & 94.57 $\pm$ 0.09                                                 \\
                                      & JOAO-v2~\cite{you2021graph} & 75.36 $\pm$ 0.47                                     & 85.96 $\pm$ 0.98                                                 & 91.15 $\pm$ 0.55                                             & 91.33 $\pm$ 0.27                                            & OOM                                                              \\
                                      & AutoGCL~\cite{yin2022autogcl} & 73.66 $\pm$ 0.59                                     & 86.44 $\pm$ 1.24                                                 & 91.98 $\pm$ 0.58                                             & 92.26 $\pm$ 0.32                                            & OOM                                                              \\
                                      & GIMM    & \textbf{79.19 $\pm$ 0.13}           & \textbf{89.29 $\pm$ 0.05}                       & \textbf{93.52 $\pm$ 0.33}                   & \textbf{93.61 $\pm$ 0.15}                  & \textbf{95.91 $\pm$ 0.10}                       \\ \bottomrule[1pt]
            \end{tabular}}
    \label{tab:node_clf}
    \vspace{0.3cm}
        \scalebox{0.59}{
        \begin{tabular}{@{}llrrrrrrrr@{}}
            \toprule[1pt]
            Type                      &        Model     & MUTAG                     & PROTEINS                  & DD                        & NCI1                      & COLLAB           & IMDB-B                    & RDT-B                  & RDT-M-5K               \\ \midrule
            w/o GDA               &InfoGraph~\cite{sun2019infograph} & 89.01 $\pm$ 1.13          & 74.44 $\pm$ 0.31          & 72.85 $\pm$ 1.78          & 76.20 $\pm$ 1.06          & 70.65 $\pm$ 1.13 & 73.03 $\pm$ 0.87          & 82.50 $\pm$ 1.42          & 53.46 $\pm$ 1.03          \\ \midrule
            w/ Manual GDA         &GraphCL~\cite{you2020graph}   & 86.80 $\pm$ 1.34          & 74.39 $\pm$ 0.45          & 78.62 $\pm$ 0.40          & 77.87 $\pm$ 0.41    & 71.36 $\pm$ 1.15 & 71.14 $\pm$ 0.44          & {\ul 89.53 $\pm$ 0.84}    &  55.99 $\pm$ 0.28    \\
                                      &MVGRL     & 89.70 $\pm$ 1.10          & -                         & -                         & -                         & -                & 74.20 $\pm$ 0.70          & 84.50 $\pm$ 0.60          & -                         \\
                                      &GCA       & {\ul 90.60 $\pm$ 0.76}    &  75.53 $\pm$ 0.22    & {\ul 79.17 $\pm$ 0.39}         & 75.87 $\pm$ 0.96          & \textbf{76.67 $\pm$ 0.43}   & {\ul 74.97 $\pm$ 0.40}          & 87.03 $\pm$ 0.90          & 56.04 $\pm$ 0.28          \\ \midrule
            w/ Automated GDA       &AD-GCL    & -                         & 73.59 $\pm$ 0.65          & 74.49 $\pm$ 0.52          & 69.67 $\pm$ 0.51          & 73.32 $\pm$ 0.61 & 71.57 $\pm$ 1.01          & 85.52 $\pm$ 0.79          & 53.00 $\pm$ 0.82          \\
                                    &JOAOv2    & -                         & 71.25 $\pm$ 0.85          & 66.91 $\pm$ 1.75          & 72.99 $\pm$ 0.75          & 70.40 $\pm$ 2.21 & 71.60 $\pm$ 0.86          & 78.35 $\pm$ 1.38          & 45.57 $\pm$ 2.86          \\
                                    &AutoGCL   & 88.64 $\pm$ 1.08          & {\ul 75.80 $\pm$ 0.36}          & 77.57 $\pm$ 0.60          & {\ul 82.00 $\pm$ 0.29}         & 70.12 $\pm$ 0.68 & 73.30 $\pm$ 0.40         & 88.58 $\pm$ 1.49          &{\ul 56.75 $\pm$ 0.18}          \\
                                &GIMM      & \textbf{91.57 $\pm$ 0.57} & \textbf{76.58 $\pm$ 0.29} & \textbf{79.32 $\pm$ 0.26} & \textbf{83.12 $\pm$ 0.20} & {\ul 76.19 $\pm$ 0.71} & \textbf{75.54 $\pm$ 0.29} & \textbf{91.29 $\pm$ 0.41} & \textbf{57.02 $\pm$ 0.14} \\ \bottomrule[1pt]
            \end{tabular}}
        \label{tab:graph_clf}
        \vspace{-0.4cm}
    \end{table}

        \begin{table*}[h]
            \centering
            \caption{Semi-supervised learning performance on graph classification (average accuracy $\pm$ std. over 5 runs).
            The highest and second performances are in \textbf{bold} and {\ul underlined} respectively.\\}
            \scalebox{0.62}{  
                \begin{tabular}{@{}lrrrrrrrrc@{}}
                    \toprule
                    Dataset     & GCA                    & GraphCL          & JOAOv2                 & AD-GCL                    & AutoGCL          & GIMM-Fit                                         & GIMM-Un-Fit-A                                    & GIMM-Un-Fit            & Ranks \\ \midrule
                    PROTEINS    & 73.85 $\pm$ 5.56       & 74.21 $\pm$ 4.50 & 73.31 $\pm$ 0.48       & 73.96 $\pm$ 0.47          & 75.65 $\pm$ 2.40 & \textbf{78.31 $\pm$ 1.31}                        & {\ul 76.46 $\pm$ 0.67}                           & 74.71 $\pm$ 0.59       & 1         \\
                    DD          & 76.74 $\pm$ 4.09       & 76.65 $\pm$ 5.12 & 75.81 $\pm$ 0.73       & 77.91 $\pm$ 0.73          & 77.50 $\pm$ 4.41 & \textbf{80.69 $\pm$ 0.47}                        & {\ul 79.30 $\pm$ 1.04}                           & 76.38 $\pm$ 0.91       & 1         \\
                    NCI1        & 68.73 $\pm$ 2.36       & 73.16 $\pm$ 2.90 & 74.86 $\pm$ 0.39       & 75.18 $\pm$ 0.31          & 73.75 $\pm$ 2.25  & {\color[HTML]{333333} \textbf{76.43 $\pm$ 0.50}} & {\ul 76.13 $\pm$ 0.59}                           & 74.32 $\pm$ 0.19       & 1         \\
                    COLLAB      & 74.32 $\pm$ 2.30       & 75.50 $\pm$ 2.15 & 75.53 $\pm$ 0.18       & 75.82 $\pm$ 0.26          & 77.16 $\pm$ 1.48 & {\color[HTML]{333333} 75.56 $\pm$ 0.30}          & {\color[HTML]{000000} \textbf{78.50 $\pm$ 0.48}} & {\ul 77.21 $\pm$ 0.47} & 1         \\
                    GITHUB      & 59.24 $\pm$ 3.21       & 63.51 $\pm$ 1.02 & {\ul 66.66 $\pm$ 0.60} & -                         & 62.46 $\pm$ 1.51 & 65.62 $\pm$ 0.15                                 & \textbf{68.46 $\pm$ 0.23}                        & 66.32 $\pm$ 0.41       & 1         \\
                    IMDB-B      & {\ul 73.70 $\pm$ 4.88} & 68.10 $\pm$ 5.15 & -                      & -                         & 71.90 $\pm$ 4.79 & \textbf{74.16 $\pm$ 0.57}                        & 71.96 $\pm$ 0.84                                 & 69.66 $\pm$ 1.00       & 1         \\
                    REDDIT-B    & 77.15 $\pm$ 6.96       & 78.05 $\pm$ 2.65 & {\ul 88.79 $\pm$ 0.65} & \textbf{90.10 $\pm$ 0.15} & 79.80 $\pm$ 3.47 & 82.68 $\pm$ 0.57                                 & 87.96 $\pm$ 0.35                                 & 85.91 $\pm$ 1.94       & 3         \\
                    REDDIT-M-5K & 32.95 $\pm$ 10.89      & 48.09 $\pm$ 1.74 & 52.71 $\pm$ 0.28       & {\ul 53.49 $\pm$ 0.28}    & 49.91 $\pm$ 2.70 & 50.16 $\pm$ 0.77                                 & {\color[HTML]{333333} \textbf{53.96 $\pm$ 0.51}} & 52.77 $\pm$ 0.57       & 1         \\ \bottomrule
                    \end{tabular}}
        \label{table:semi}
        \vspace{-0.5cm}
            \end{table*}

            \begin{table*}[h]
                \centering
                \caption{Ablation studies on GIMM (average accuracy $\pm$ std. over 3 runs). Different ablation groups are separated by different background colors. The highest performance on the first and the second ablation group are {\ul underlined} and in \textbf{bold} respectively.\\}
                \scalebox{0.62}{    
                    \begin{tabular}{@{}l
                        >{\columncolor[HTML]{FFFFFF}}r 
                        >{\columncolor[HTML]{FFFFFF}}r 
                        >{\columncolor[HTML]{FFFFFF}}r 
                        >{\columncolor[HTML]{EFEFEF}}r 
                        >{\columncolor[HTML]{EFEFEF}}r 
                        >{\columncolor[HTML]{EFEFEF}}r r@{}}
                        \toprule[1pt]
                        Dataset          & GIMM-Uni                                        & GIMM-Feat                & GIMM-Edge                                        & GIMM-Simp                 & GIMM-ViewM                                       & GIMM-Simult               & GIMM                            \\ \midrule
                        Wiki-CS          & {{\ul 79.30 $\pm$ 0.00}} & 76.43 $\pm$ 0.15       & 79.16 $\pm$ 0.07                              & 79.10 $\pm$ 0.15          & {\textbf{79.34 $\pm$ 0.05}} & 79.13 $\pm$ 0.12          & 79.29 $\pm$ 0.05                \\
                        Amazon-Computers & 88.12 $\pm$ 0.25                              & 86.74 $\pm$ 0.28       & 88.87 $\pm$ 0.17                              & 89.12 $\pm$ 0.16          & 89.04 $\pm$ 0.08                                 & 89.00 $\pm$ 0.32          & {\ul \textbf{89.29 $\pm$ 0.08}} \\
                        Amazon-Photo     & 93.27 $\pm$ 0.47                              & 91.21 $\pm$ 0.49       & 93.57 $\pm$ 0.26                              & 93.65 $\pm$ 0.32          & 93.44 $\pm$ 0.39                                 & \textbf{93.80 $\pm$ 0.27} & {\ul 93.68 $\pm$ 0.35}          \\
                        Coauthor-CS      & 93.32 $\pm$ 0.05                              & 93.41 $\pm$ 0.10       & 93.33 $\pm$ 0.02                              & 93.44 $\pm$ 0.08          & 93.48 $\pm$ 0.02                                 & 93.39 $\pm$ 0.12          & {\ul \textbf{93.62 $\pm$ 0.10}} \\ \midrule
                        MUTUG            & 90.79 $\pm$ 0.28                              & 91.13 $\pm$ 0.77       & 90.24 $\pm$ 0.42                              & 91.53 $\pm$ 1.02          & 90.45 $\pm$ 0.54                                 & 91.35 $\pm$ 0.61          & {\ul \textbf{91.83 $\pm$ 0.62}} \\
                        PROTEINS         & {76.28 $\pm$ 0.09}       & 75.17 $\pm$ 0.37       & 76.25 $\pm$ 0.51                              & 76.49 $\pm$ 0.19          & 76.53 $\pm$ 0.28                                 & \textbf{77.30 $\pm$ 0.21} & {\ul 76.61 $\pm$ 0.37}          \\
                        DD               & 79.37 $\pm$ 0.59                              & 76.62 $\pm$ 0.48       & {{\ul 79.62 $\pm$ 0.09}} & 79.20 $\pm$ 0.22          & 79.00 $\pm$ 0.38                                 & 78.83 $\pm$ 0.05          & \textbf{79.43 $\pm$ 0.34}       \\
                        COLLAB           & 72.30 $\pm$ 1.06                              & 74.00 $\pm$ 0.28       & {{\ul 77.02 $\pm$ 1.46}} & 70.39 $\pm$ 0.95          & 75.91 $\pm$ 0.10                                 & 74.12 $\pm$ 0.63          & \textbf{76.08 $\pm$ 0.86}       \\
                        IMDB-B           & 75.20 $\pm$ 0.70                              & 72.67 $\pm$ 0.06       & 75.07 $\pm$ 0.76                              & 70.23 $\pm$ 1.50          & 75.33 $\pm$ 1.07                                 & 75.60 $\pm$ 0.66          & {\ul \textbf{75.70 $\pm$ 0.26}} \\
                        REDDIT-B         & 90.05 $\pm$ 0.61                              & {\ul 92.00 $\pm$ 0.48} & 90.05 $\pm$ 0.30                              & \textbf{91.63 $\pm$ 0.25} & 90.90 $\pm$ 0.38                                 & 89.92 $\pm$ 0.40          & 91.25 $\pm$ 0.75                \\ \bottomrule[1pt]
                        \end{tabular}}
                    \label{tab:ablation}
                    \vspace{-0.3cm}
                    \end{table*}

The feature matrix of $\mathcal{G}_1$ can be represented by $\textbf{X}_1=\textbf{X}\odot \textbf{M}$, where $\textbf{M}\in \mathbb{R}^{N\times F}$ is broadcast from $ \textbf{m}\in \{0,1\}^{F}$.
A random variable $p_{f_k}\sim \operatorname{Bernoulli}(1-p_{d,f_k})$ is used to select the $k$-th dimension feature, i.e., if $p_{f_k}=1$, then $\textbf{m}_k=1$, else $\textbf{m}_k=0$. $\textbf{m}_k$ represents the $k$-th dimension of $\textbf{m}$.
$\textbf{P}_{d,f}=[p_{d,f_1},...,p_{d,f_F}]$ is given by
$
    \textbf{P}_{d,f} = \min(((\textbf{P}_f^{\max}-\textbf{P}_f)/(\textbf{P}_f^{\max}-\textbf{P}_f^{\operatorname{avg}}))\cdot p_{s2},p_t),
$
where $p_{s2}$ is a hyperparameter between 0 and 1 to control the overall mask rate of features.

\subsection{View comparison module}
InfoMax can be applied to these approximate optimal views without the interference of nuisance information or the loss of noteworthy information. 
The node classification task aims to learn an encoder $f_n: \mathcal{G}\rightarrow\mathbb{R}^{N\times d}$. $\mathcal{G}_1$ and $\mathcal{G}_2$ are the views of $\mathcal{G}$, then our objective is
\begin{equation}
    \setlength\abovedisplayskip{2pt}
\setlength\belowdisplayskip{-3pt}
    \max_{f_n,g_n}I(g_n(f_n(\mathcal{G}_1));g_n(f_n(\mathcal{G}_2))),
\end{equation}
   
where $f_n$ is a GCN~\cite{welling2016semi}, $g_n$ is a simple 1 or 2- layer MLP, and we use InfoNCE (Eqn. \ref{equa:infonce}) as the estimator of mutual information.
The graph classification task aims to learn an encoder $f_g: \mathcal{G}_i \rightarrow \mathbb{R}^{d}$. Given a minibatch of graphs $\{\mathcal{G}_i\}_{i=1}^K$,
$\mathcal{G}_{i,1}$ and $\mathcal{G}_{i,2}$ are the views of $\mathcal{G}_i$, then the objective is
\begin{equation}
    \setlength\abovedisplayskip{2pt}
    \setlength\belowdisplayskip{-3pt}
    \max_{f_g,g_g}\frac{1}{K}\sum_{i=1}^{K}I(g_g(f_g(\mathcal{G}_{i,1}));g_g(f_g(\mathcal{G}_{i,2}))),
\end{equation}

where $f_g$ is a GIN~\cite{xu2018powerful}, $g_g$ is a simple 1 or 2- layer MLP followed by a readout function which is a simple summation. 
$f_n$ and $f_g$ can be replaced by any graph encoder. 
Finally, the node embeddings $f_n(\mathcal{G})$ and graph embeddings $\{f_g(\mathcal{G}_i), i = 1,...,Q\}$ are used for downstream testing.
More details of the algorithm and the computational complexity analysis can be found in Appendix.

\section{Experiments}\label{exp}
\textbf{Dataset details.}
We use 5 datasets for node classification, including Wiki-CS\footnote{\url{https://github.com/pmernyei/wiki-cs-dataset/raw/master/dataset}}~\cite{mernyei2020wiki}, Amazon-Computers\footnote{\url{https://github.com/shchur/gnn-benchmark/raw/master/data/npz/amazon\_electronics\_computers.npz}}, Amazon-Photo\footnote{\url{https://github.com/shchur/gnn-benchmark/raw/master/data/npz/amazon\_electronics\_photo.npz}}, Coauthor-CS\footnote{\url{https://github.com/shchur/gnn-benchmark/raw/master/data/npz/ms\_academic\_cs.npz}} and Coauthor-Physics\footnote{\url{https://github.com/shchur/gnn-benchmark/raw/master/data/npz/ms\_academic\_phy.npz}}~\cite{shchur2018pitfalls}.
Their detailed statistics are shown in Table \ref{tab:node_clf_data}.
Wiki-CS has dense real number features, whereas the other datasets have sparse one-hot features.
Following GCA~\cite{zhu2021graph}, we evaluate models under the public train, test, and validation sets supplied by Wiki-CS. 
For the other four datasets, we randomly split the dataset into three sets: 80\% train, 10\% test, and 10\% validation.
We use 9 datasets from TUDataset\footnote{\url{https://chrsmrrs.github.io/datasets/docs/datasets/}}~\cite{morris2020tudataset} for graph classification, including MUTUG, PROTEINS, DD, NCI1, COLLAB, GITHUB, IMDB-BINARY, REDDIT-BINARY, and REDDIT-MULTI-5K.
Their detailed statistics are shown in Table \ref{tab:graph_clf_data}.
Following JOAO~\cite{you2021graph}, we use the entire dataset to learn graph representations and feed them into the downstream classifier using 10-fold cross-validation.

\textbf{Experimental setup.}
For node classification, GIMM is compared to manual data augmentation GCL methods (MVGRL~\cite{hassani2020contrastive}, GCA~\cite{zhu2021graph}), automated data augmentation GCL methods (AD-GCL~\cite{suresh2021adversarial}, JOAOv2~\cite{you2021graph}, AutoGCL~\cite{yin2022autogcl}), and GCL methods without data augmentation (DGI~\cite{velickovic2019deep}, GMI~\cite{peng2020graph}).
For graph classification, GIMM is compared to manual data augmentation GCL methods (GraphCL~\cite{you2020graph}, MVGRL, GCA), automated data augmentation GCL methods (AD-GCL, JOAOv2, AutoGCL), and GCL method without data augmentation (InfoGraph~\cite{sun2019infograph}).
The results of AD-GCL, JOAOv2, and AutoGCL on node classification and GCA on graph classification are not provided in the original papers; hence, we remove or add a pooling layer to get the node and graph representations.
We report the best results for GCA's 3 variations, GCA-DE, GCA-EVC, and GCA-PR.

We train GIMM and other baselines using unlabeled data to generate representations, then train classifiers for downstream tasks using these representations.
The downstream classifier for node classification is a $\ell_2$-regularized logistic regression with a learning rate of 0.01.
The downstream classifier for graph classification is an SVM with parameter $C$ grid searching in [0.001, 0.01, 0.1, 1, 10, 100, 1000].
On all datasets, the Xavier initialization~\cite{glorot2010understanding} and Adam optimizer~\cite{kingma2014adam} are utilized. 
In GIMM, the sum of node representations of the two views and the original graph: $f_n(\mathcal{G}_1)+f_n(\mathcal{G}_2)+2*f_n(\mathcal{G})$ or the sum of node representations of the two views: $f_n(\mathcal{G}_1)+f_n(\mathcal{G}_2)$ is used for node classification, considering that incorporating view representations can lead to more generalized representations. 
Graph representations $\{f_g(\mathcal{G}_i),i=1,...,Q\}$ are used for graph classification.

\subsection{Comparison with the state-of-the-art methods}
In this section, we compare GIMM with the SOTA methods in unsupervised and semi-supervised learning settings for node and graph classification.

\textbf{Unsupervised learning on node classification.} 
Table \ref{tab:node_clf} (TOP) shows that GIMM achieves SOTA performance on the node classification task. 
Firstly, it significantly outperforms baselines with automated data augmentation.
The reason for the improvement is that the views, defined by retaining minimal noteworthy information, approximate the optimal views in InfoMin while other baselines fail to.
Utilizing such views can prevent the contrastive module from learning nuisance information and make it focuses on noteworthy information, thus achieving better performance.
Secondly, GIMM outperforms manual augmentation baselines on all datasets.
GCL methods with manual augmentation heavily rely on the predefined data augmentation pool, resulting in a limited approximation of the InfoMin principle.
Considering factors such as time cost, computational cost, and the ability to approximate optimal views in InfoMin, GIMM emerges as a preferable alternative to GCL with manual GDA.
Thus, GIMM liberates GCL from the tedious manual selection of data augmentation. 

\textbf{Unsupervised learning on graph classification.} 
Table \ref{tab:graph_clf} (BOTTOM) shows that GIMM achieves the best performance on 7 out of 8 datasets and surpasses automated data augmentation methods on all datasets.
The results of the graph and node classification tasks prove that our strategy of using minimal noteworthy information to approximate minimal necessary information is effective on different downstream tasks.
GIMM approximates the optimal views of InfoMin, without requiring task-relevant information.
Lastly, Table \ref{tab:node_clf} (TOP and BOTTOM) illustrates that methods with data augmentation generally outperform methods without data augmentation, highlighting the significance of data augmentation in GCL.

\textbf{Semi-supervised learning on graph classification.}
Following the experimental setup in AutoGCL, we perform semi-supervised learning on TUDataset for graph classification using 10-fold cross-validation.
GIMM-Fit is trained and tested using 10\% labeled data.
GIMM-Un-Fit-A involves several alternate training steps, where each step comprises training with 80\% unlabeled data and fine-tuning with 10\% labeled data. Finally, it is tested on 10\% labeled data.
GIMM-Un-Fit is trained on 80\% unlabeled data, fine-tuned on 10\% labeled data, and tested on 10\% labeled data.
Table \ref{table:semi} shows that GIMM outperforms baselines on 7 out of 8 datasets and gets an average rank of 1.25.
Intriguingly, unlabeled data is not always effective, and the performance of GIMM-Un-Fit is sometimes worse than GIMM-Fit trained with only 10\% labeled data.
We speculate that using labels directly to fine-tune GIMM trained with unlabeled data will interfere with the distribution already learned, resulting in a performance loss.
GIMM-Un-Fit-A outperforms GIMM-Un-Fit by using unlabeled and labeled data alternately.
The alternating training approach aids in mitigating the divergence of learned knowledge between unlabeled and labeled data, thereby fostering mutual learning between the two sources.

\subsection{Ablation studies}
Ablation studies on 10 datasets, conducted with identical hyperparameters (including random seeds), are presented in Table \ref{tab:ablation} to validate the rationale behind the components of GIMM.

\textbf{Effectiveness of automated view generator.}
We use 3 variants to verify the effectiveness of the generator.
GIMM-Uni employs uniformly distributed edge and feature importance.
GIMM-Feat employs uniformly distributed edge importance but retains the automated generation of feature importance.
GIMM-Edge employs uniformly distributed feature importance but retains the automated generation of edge importance.
GIMM outperforms GIMM-Uni on all datasets except Wiki-CS, and performs similarly on Wiki-CS, demonstrating the effectiveness of the automated view generator.
GIMM outperforms GIMM-Feat on 9 datasets and outperforms GIMM-Edge on 8 datasets.
The results indicate that combining edge and feature importance is more effective than using only one. Therefore, GDA should consider both topology and features to achieve optimal performance.

\textbf{Rationality of max-min optimization.}
We employ 3 variations to validate the rationality of adversarial optimization in the automated view generator.
   GIMM-Simp replaces the learnable GCN encoder in the view generator with a parameterless GCN, i.e., $\textbf{W}=\textbf{I}$.
   GIMM-ViewM optimizes the view generator through max-max optimization.
   \begin{figure}[t]
    \centering
        \subfigbottomskip=0pt
     {\includegraphics[scale=0.5]{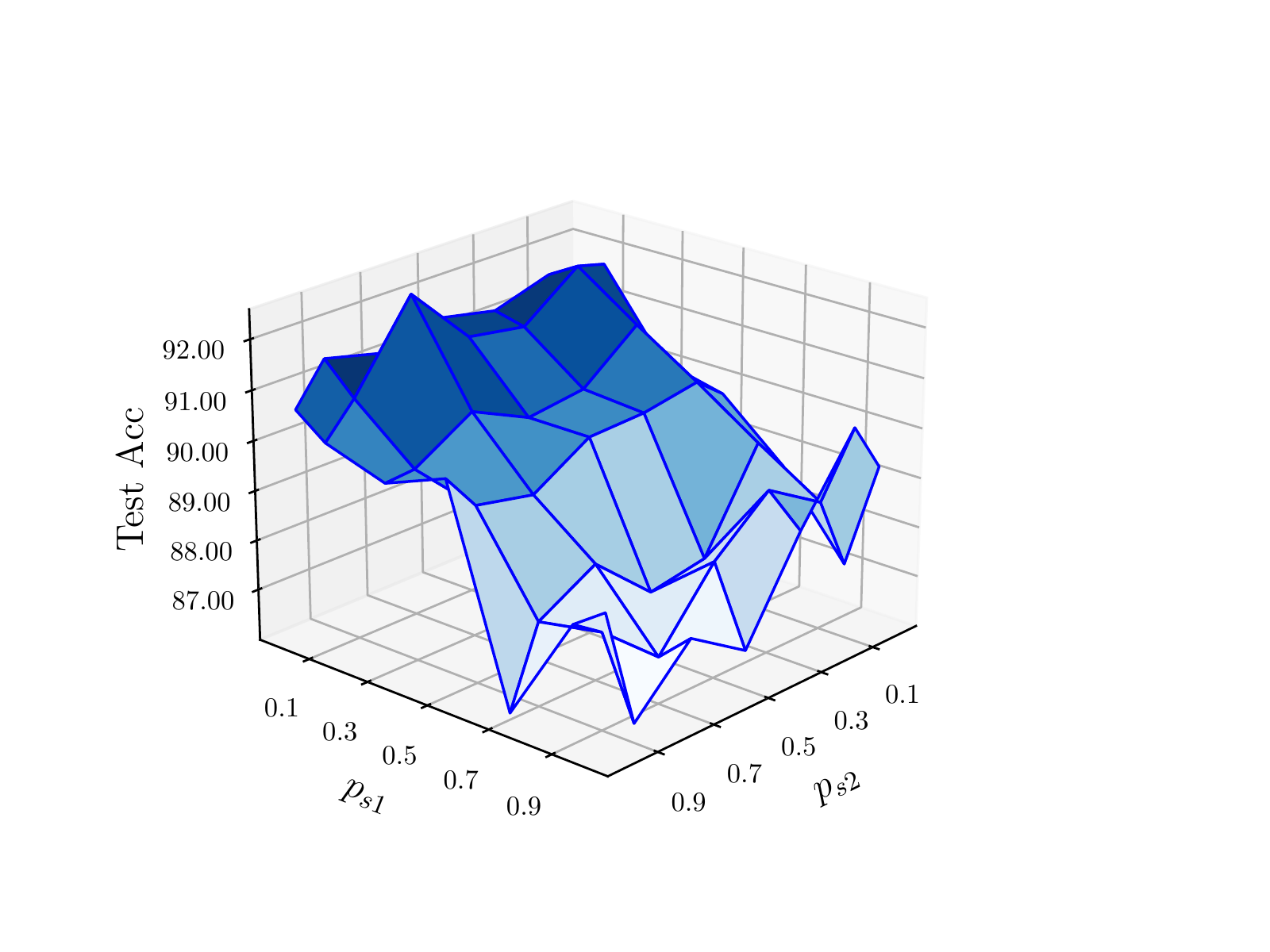}}
     {\includegraphics[scale=0.5]{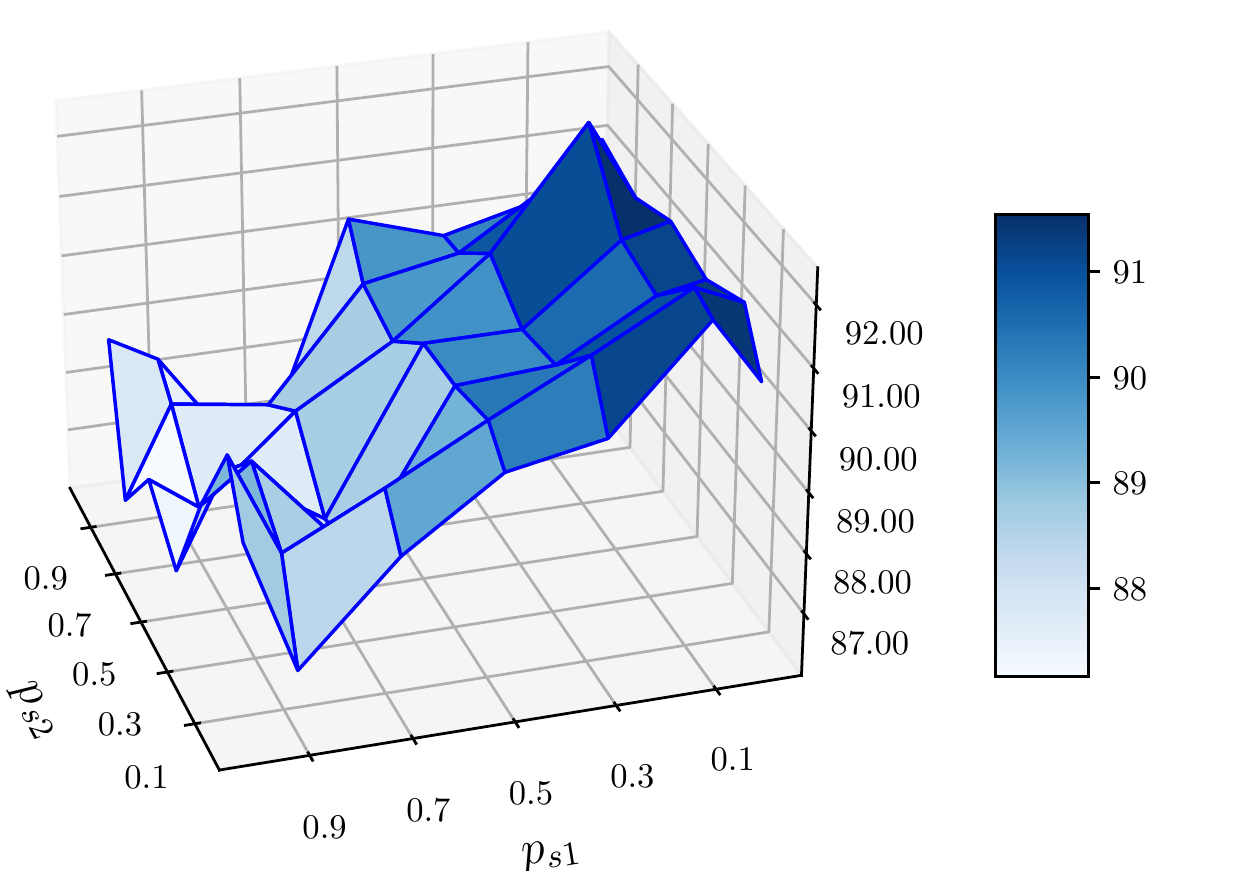}}
     \caption{Graph classification performance of GIMM with different $p_{s1}$ and $p_{s2}$ on REDDIT-BINARY.}
     \label{fig:sensity}
 \end{figure}
   \begin{figure}[t]
    \centering
    \vspace{-0.3cm}
    \setlength{\belowcaptionskip}{-0.3cm}
    \subfigure[Edge importance visualization.]{\includegraphics[scale=0.6]{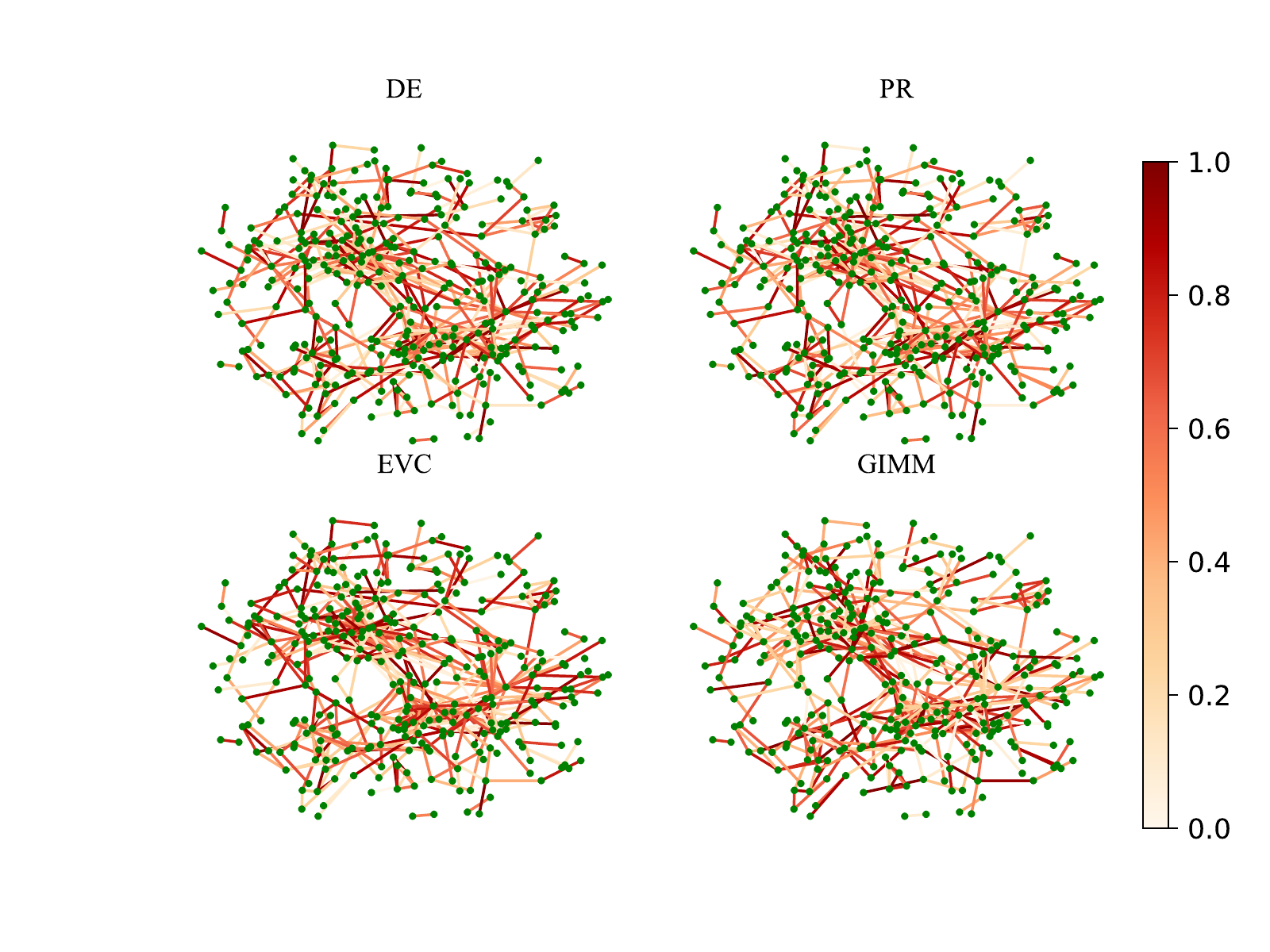}\label{fig:graph_map}}
    \subfigure[Feature importance visualization. 50 dimensions from the features are selected at random for display.]{\includegraphics[scale=0.6]{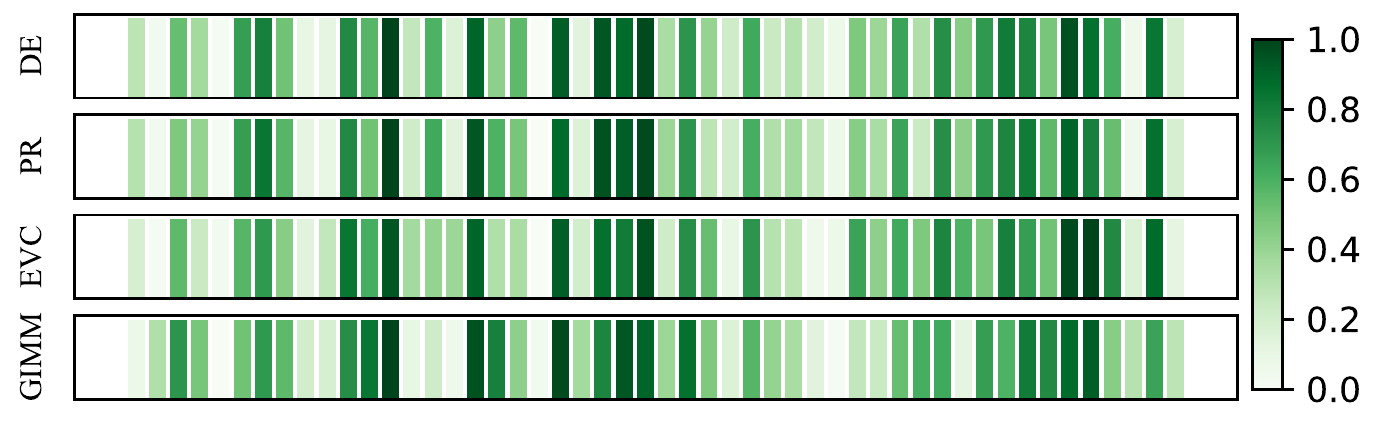}\label{fig:feature_map}}
    \caption{DE, PR, and EVC are degree, PageRank and eigenvector centrality respectively.}
    \label{fig: figure}
\end{figure}
\begin{figure}[t]
  \centering
 \subfigbottomskip=1pt 
  \subfigcapskip=-5pt 
      {\includegraphics[width=0.18\linewidth]{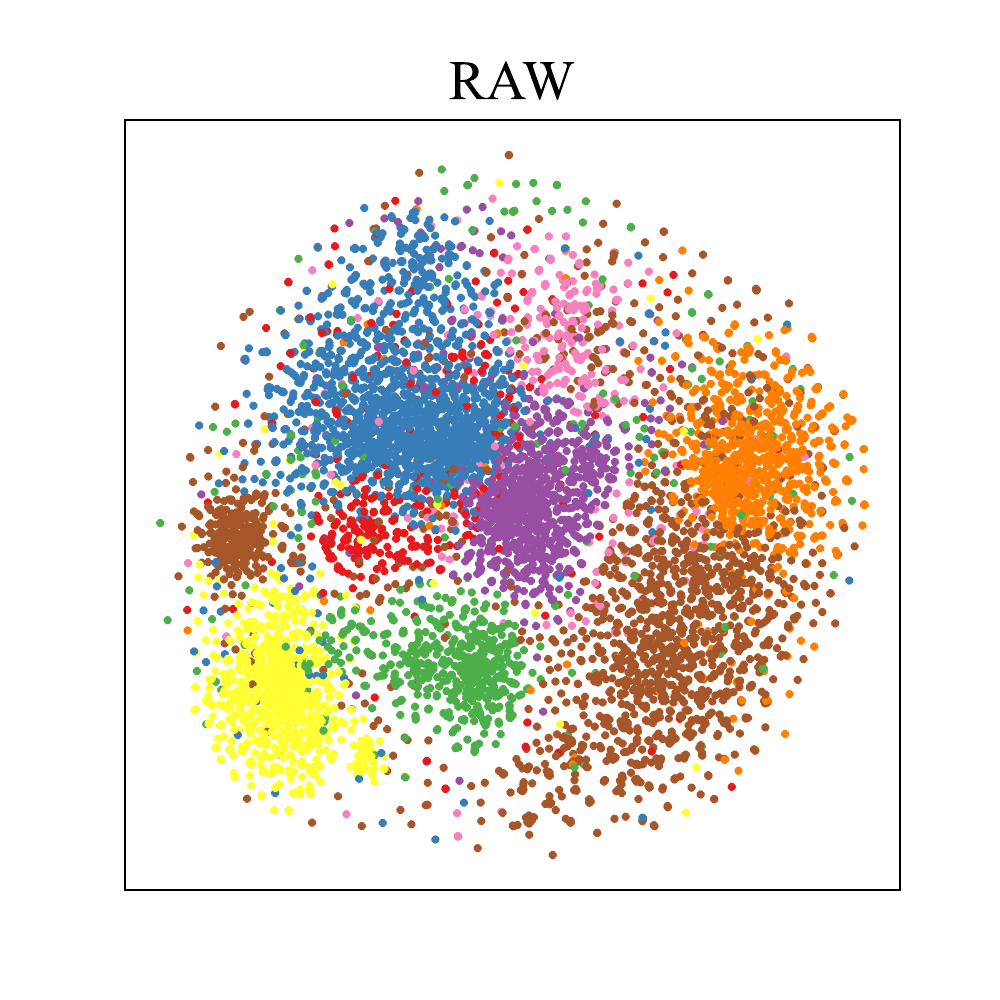}} \hspace{-0.005\linewidth}
      {\includegraphics[width=0.18\linewidth]{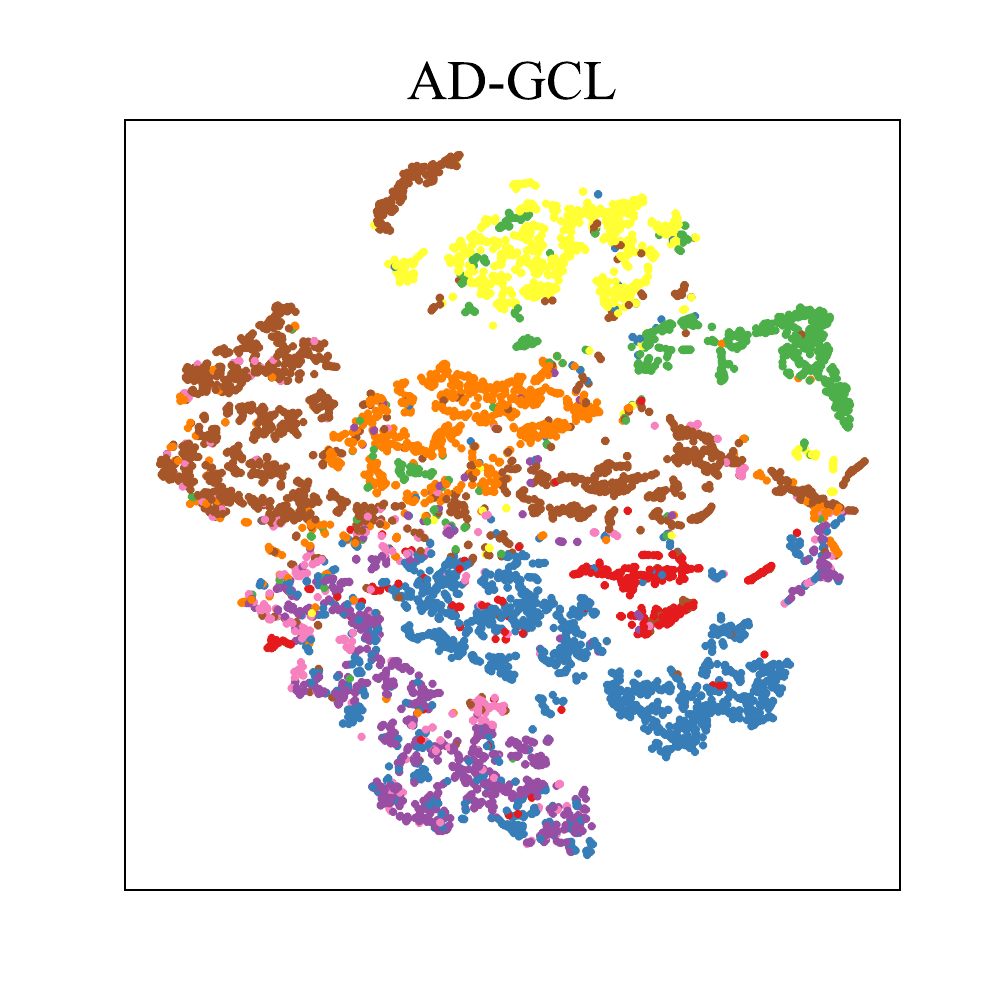}} \hspace{-0.005\linewidth}
      {\includegraphics[width=0.18\linewidth]{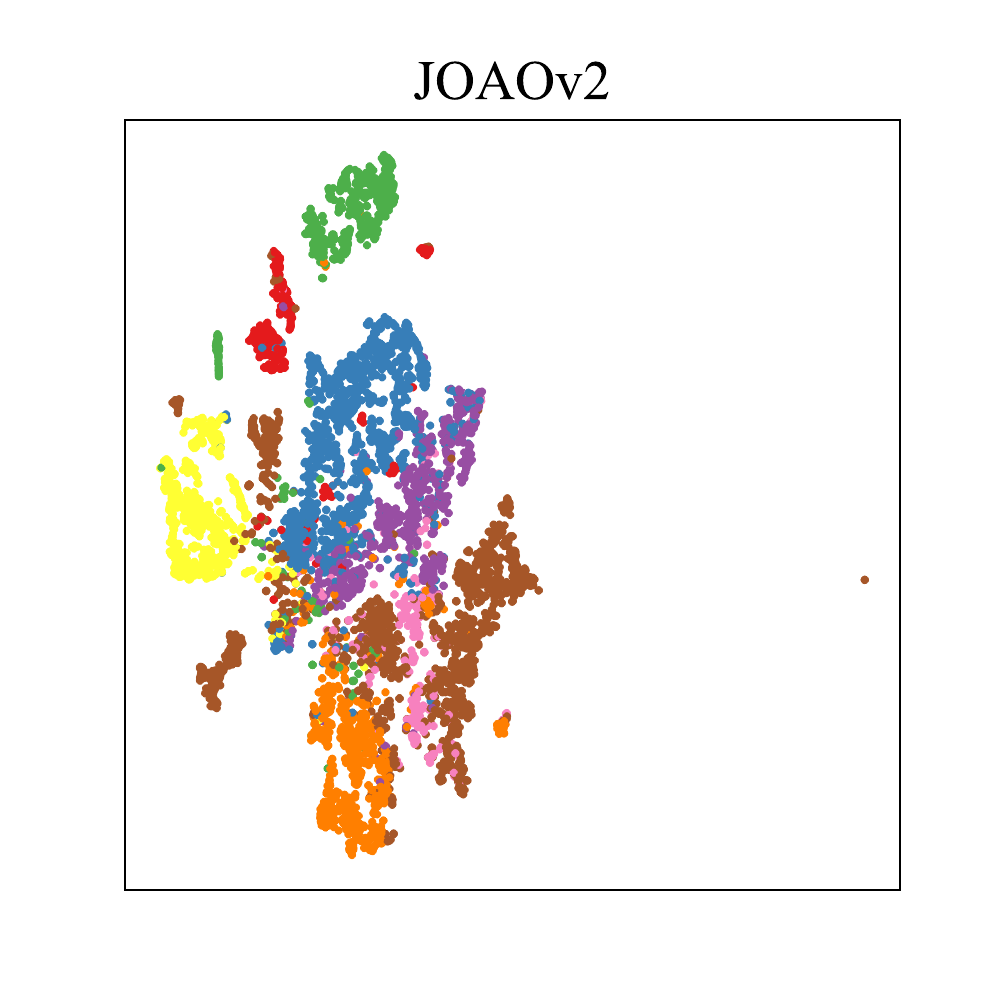}}\hspace{0.005\linewidth}
      {\includegraphics[width=0.18\linewidth]{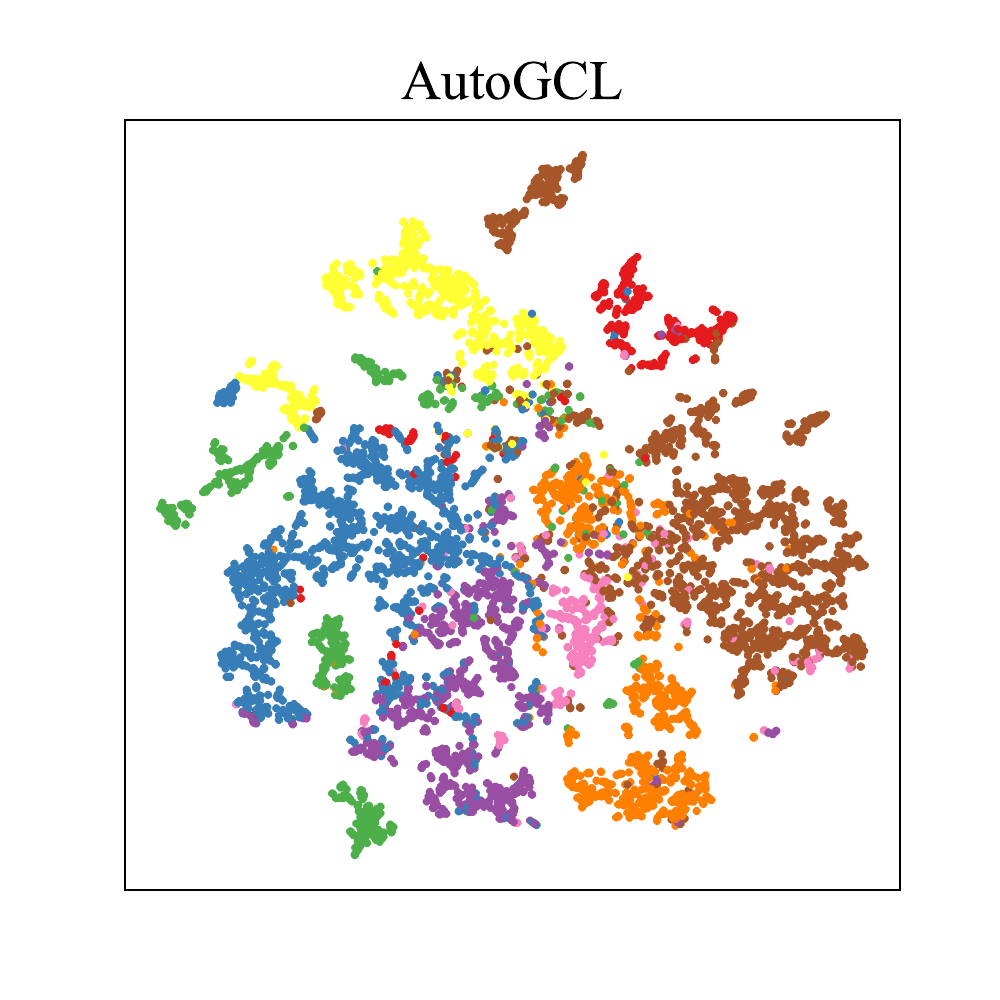}} \hspace{-0.005\linewidth}
      {\includegraphics[width=0.18\linewidth]{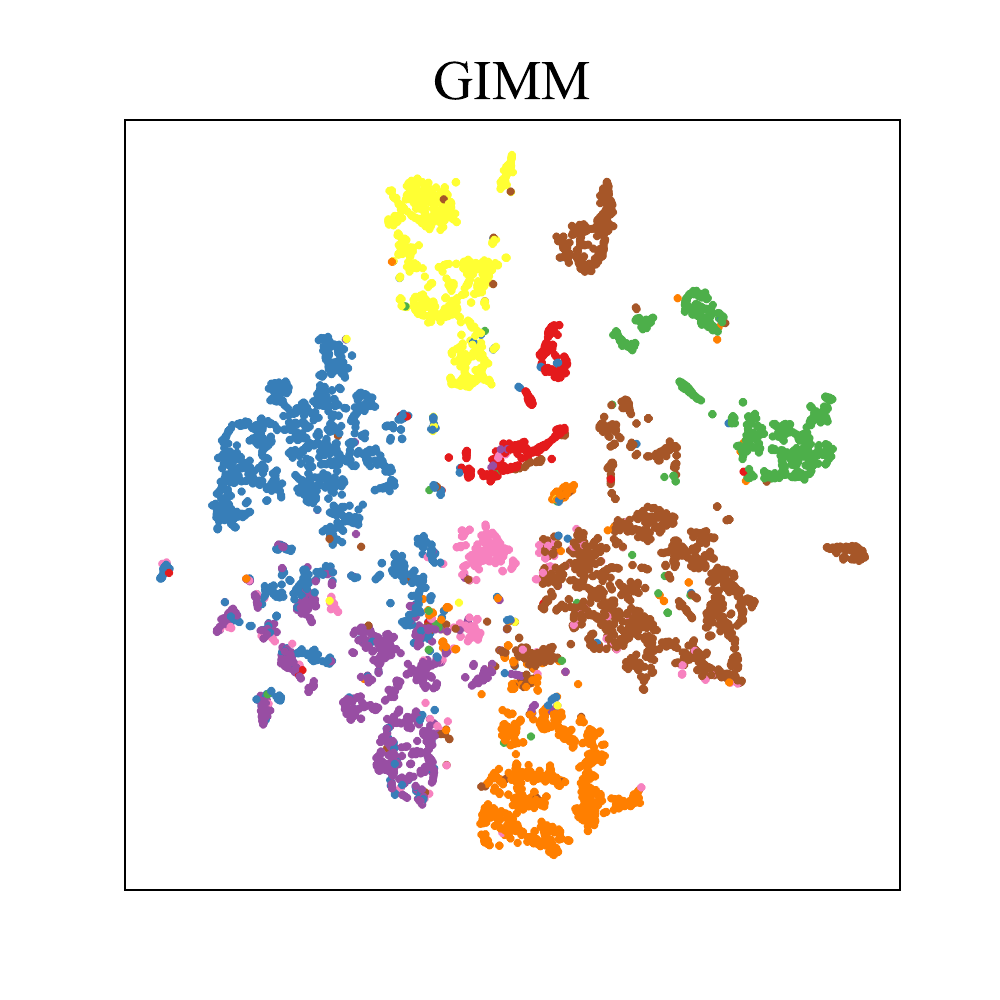}}
  \caption{t-SNE visualization of raw features and node representations on Amazon-Photo. Different colors represent different node classes.}
  \label{fig:tsne}
\end{figure}
   GIMM-Simult trains the view generator and comparison module simultaneously, using the GNN encoder from the view comparison module as the GNN encoder for the view generator.
Firstly, GIMM performs better than GIMM-Simp on 9 datasets, proving that the learnable GCN encoder in the generator is effective. 
Secondly, GIMM outperforms GIMM-ViewM on 9 datasets, showing that the GNN encoder and the projection head trained by minimizing are better than maximizing. 
Minimizing the objective yields an aggressive graph encoder and projection head that contribute to generating a more generalized importance graph and a more precise approximation of InfoMin's optimal views.
 Finally, GIMM beats GIMM-Simult on 8 datasets, as training the view generator and the comparison module simultaneously prevents the generation of superior views and representations.
The view generator emphasizes remaining noteworthy information and discarding nuisance information, whereas the comparison module highlights the quality of the graph encoder.
They impede each other when training together.

\textbf{Sensitivity analysis.}
We conduct sensitivity analysis on critical hyperparameters $p_{s1},p_{s2}$ of GIMM.
To simplify the analysis, $p_{s1}$ and $p_{s2}$ of the two views are set to be identical, and they are utilized to adjust the overall drop rate of edges and the overall mask rate of features.
$p_{s1}$ and $p_{s2}$ are selected from [0.0, 0.1, 0.3, 0.7, 0.9, 1.0].
The results are shown in Figure \ref{fig:sensity}.

Firstly, it can be observed that GIMM is more sensitive to $p_{s1}$ than $p_{s2}$; altering $p_{s1}$ results in a more noticeable performance change. 
This observation suggests that topology information plays a more crucial role in GIMM than feature information.
Secondly, it is worth noting that excessively large values of $p_{s1}$ or $p_{s2}$ have a detrimental effect on performance. 
When $p_{s1}$ becomes excessively large, the topology information of the graph is almost completely destroyed, resulting in isolated nodes that lack connections.
When $p_{s2}$ becomes excessively large, the graph almost degenerates into a featureless graph.
The performance tends to be extremely poor when both are large.
Thirdly, points in Figure \ref{fig:sensity} surrounding (0,0) perform better than (0,0).
This observation suggests that utilizing importance to generate views, even with small drop or mask rates, is more advantageous than solely employing the original graph for comparison.

\textbf{Visualisation.} Centrality reflects the significance of a node in a graph.
In GCA, edge and feature importance are defined by degree, PageRank, and eigenvector centrality.
To demonstrate the reasonableness of the importance gained by GIMM, we visualize the importance defined by those centralities and obtained by GIMM on a random subgraph of Amazon-Photo.
According to Figure \ref{fig:graph_map}, the edge importances highlight two distinct groups situated in the upper left and lower right corners of the graph. Compared with the importance defined by centrality, GIMM pays more attention to the backbone edges within the group and the edges associated with each group, which is the fundamental topology of the graph.
In Figure \ref{fig:feature_map}, the importance features gained by GIMM are roughly aligned with those acquired by centrality, but GIMM emphasizes fewer features.

To demonstrate the quality of the representations, 
we employ two components PCA~\cite{abdi2010principal} and t-SNE~\cite{van2008visualizing} to visualize the raw features and representations of AD-GCL, JOAOv2, AutoGCL, and GIMM on Amazon-Photo.
Figure \ref{fig:tsne} demonstrates that GIMM achieves a more distinct classification boundary, showcasing the superior representations learned by GIMM. 

\section{Conclusion}
In this paper, we propose GIMM, a novel GCL method featuring automated data augmentation.
 To the best of our knowledge, this is the first method that combines InfoMin and InfoMax in GCL.
 GIMM approximates InfoMin's optimal views by replacing minimal necessary information with minimal noteworthy information, without requiring task-relevant information. 
 Applying InfoMax to these views can avoid the risk of redundant information and insufficient information.
 In addition, GIMM introduces randomness to augmentation, thus stabilizing the model against perturbations.
 Extensive experiments on node and graph classification tasks demonstrate GIMM's superiority, which outperforms both automated and manual data augmentation GCL methods.
 In the future, we will validate the effectiveness of our approximate optimal views on more downstream tasks.
\section{Appendix}
\begin{algorithm}[!h]
    \caption{Automated graph view generator}
    \label{alg:1}
    \begin{algorithmic}[1]
\REQUIRE 
graph $\mathcal{G}=(\textbf{A},\textbf{X})$; edge set $\mathcal{E}$, $|\mathcal{E}|=M$; feature dimension $F$; GCN encoder $f_\textbf{W}(\cdot)$; feature importance MLP $h_\phi(\cdot)$; edge importance MLP $h_\psi(\cdot)$; Gumbel-Max function $\operatorname{Gumbel}(\cdot)$; 
GNN encoder $f_{\theta}(\cdot)$; projection head $g_{\xi}(\cdot)$;
mutual information estimator $\hat{I}(\cdot;\cdot)$.\\  
\renewcommand{\algorithmicrequire}{ \textbf{HyperParas:}}
\REQUIRE 
regularization weight $\lambda$; scale parameters for view 1 and view 2 $p_{1,s1},p_{2,s1},p_{1,s2},p_{2,s2}$; truncate parameter $p_t$; learning rate $\alpha$.
\ENSURE 
    view 1 $\mathcal{G}_1 = (\textbf{A}_1,\textbf{X}_1)$; view 2 $\mathcal{G}_2 = (\textbf{A}_2,\textbf{X}_2)$.
        \FOR{ep in \#epochs}
            \STATE $\textbf{E}=f_\textbf{W}(\textbf{A},\textbf{X})$;\\
            \STATE $\textbf{P}_n = \operatorname{Gumbel}(h_\phi(\textbf{E}))$;\\
            \STATE $\textbf{P}_f=\textbf{X}^T\textbf{P}_n;$
            \FOR{ $\forall \varepsilon_k \in \mathcal{E}$}
            \STATE $\textbf{p}_{e,k} = \operatorname{Gumbel}(h_\psi([\textbf{E}[v_i];\textbf{E}[v_j]])),\varepsilon_k=(v_i,v_j)$;\\
            \ENDFOR
            \STATE $\textbf{P}_e=[\textbf{p}_{e,1},...,\textbf{p}_{e,M}]^T$;
            \STATE Broadcast $\textbf{P}_f\in\mathbb{R}^{F\times 1}$ to $\textbf{P}'_f\in\mathbb{R}^{N\times F}$;
            \STATE $\tilde{\textbf{X}} = \textbf{X}\odot\textbf{P}'_f$;
            \STATE Derive $\tilde{\textbf{A}}$ via $\textbf{P}_e$;
            \STATE $\tilde{\mathcal{G}}=(\tilde{\textbf{A}},\tilde{\textbf{X}})$;
            \STATE $\mathcal{L}=-\hat{I}(g_\xi(f_\theta(\mathcal{G}));g_\xi(f_\theta(\tilde{\mathcal{G}})))$;
            \STATE $\mathcal{R}=\frac{\sum_{i=1}^F \textbf{P}_f^{(i)}}{F}+\frac{\sum_{i=1}^M \textbf{P}_e^{(i)}}{M}$;
            \\ /* Update the importance learner */
            \STATE $\textbf{W} \leftarrow \textbf{W} -\alpha\nabla_\textbf{W} (\mathcal{L}+\lambda * \mathcal{R}) $;
            \STATE $\phi \leftarrow \phi -\alpha\nabla_\phi (\mathcal{L}+\lambda * \mathcal{R}) $;
            \STATE $\psi \leftarrow \psi -\alpha\nabla_\psi (\mathcal{L}+\lambda * \mathcal{R}) $;
            \\ /* Update the GNN encoder and projection head */
            \STATE $\theta \leftarrow \theta +\alpha\nabla_\theta (\mathcal{L}) $;
            \STATE $\xi \leftarrow \xi +\alpha\nabla_\xi (\mathcal{L}) $;
        \ENDFOR
        \STATE 
        \begin{equation}
        \scalebox{0.95}{$
            \begin{aligned}
                \textbf{P}_{d,\varepsilon}=[p_{d,\varepsilon_1},...,p_{d,\varepsilon_M}]=\min\left(\frac{\textbf{P}_{e}^{\max}-\textbf{P}_{e}}{\textbf{P}_{e}^{\max}-\textbf{P}_{e}^{\operatorname{avg}}}\cdot p_{1,s1},p_t\right); \nonumber
            \end{aligned} $}
        \end{equation}
        \STATE Sample $p_{\varepsilon_k}$ from $\operatorname{Bernoulli}(1-p_{d,\varepsilon_k}),k=1,...,M$; 
        \STATE Use $p_{\varepsilon_k},k=1,...,M$ to generate $\textbf{A}_1$;

        \STATE \begin{equation}
            \scalebox{0.95}{$
            \begin{aligned}
                \textbf{P}_{d,f}=[p_{d,f_1},...,p_{d,f_F}]=\min\left(\frac{\textbf{P}_{f}^{\max}-\textbf{P}_{f}}{\textbf{P}_{f}^{\max}-\textbf{P}_{f}^{\operatorname{avg}}}\cdot p_{1,s2},p_t\right); \nonumber
            \end{aligned}$}
        \end{equation}
        \STATE Sample $p_{f_k}$ from $\operatorname{Bernoulli}(1-p_{d,f_k}),k=1,...,F$;
        \STATE Use $p_{f_k},k=1,...,F$ to generate $\textbf{X}_1$;
        \STATE Generate $\textbf{A}_2,\textbf{X}_2$ similarly;
\RETURN $\mathcal{G}_1 = (\textbf{A}_1,\textbf{X}_1)$; $\mathcal{G}_2 = (\textbf{A}_2,\textbf{X}_2)$; 
    \end{algorithmic}
\end{algorithm}

\begin{algorithm}[!h]
    \caption{View comparison module for node classification}
    \label{alg:2}
    \begin{algorithmic}[1]
        \renewcommand{\algorithmicrequire}{ \textbf{Input:}}
        \REQUIRE 
        graph $\mathcal{G}$; GNN encoder $f_\omega(\cdot)$; projection head $g_\rho(\cdot)$; mutual information estimator $\hat{I}(\cdot;\cdot)$.\\ 
        \renewcommand{\algorithmicrequire}{ \textbf{HyperParas:}}
        \REQUIRE 
        learning rate $\beta$. 
        \ENSURE 
        Trained $f_\omega(\cdot)$.\\
        \STATE $(\mathcal{G}_1,\mathcal{G}_2)=\operatorname{Algorithm}$ \ref{alg:1} $(\mathcal{G})$;
        \FOR{ep in \#epochs}
        \STATE $\mathcal{L}=-\hat{I}(g_\rho(f_\omega(\mathcal{G}_1));g_\rho(f_\omega(\mathcal{G}_2)))$;
        \STATE $\omega \leftarrow \omega - \beta\nabla_{\omega}\mathcal{L}$; $\rho \leftarrow \rho - \beta\nabla_{\rho}\mathcal{L}$;
        \ENDFOR
\RETURN Trained $f_\omega(\cdot)$;
    \end{algorithmic}
\end{algorithm}

\begin{algorithm}[!h]
    \caption{View comparison module for graph classification}
    \label{alg:3}
    \begin{algorithmic}[1]
\renewcommand{\algorithmicrequire}{ \textbf{Input:}}
\REQUIRE 
graph set $\{\mathcal{G}_{i}\},i=1,...,Q$; GNN encoder $f_\omega(\cdot)$; projection head $g_\rho(\cdot)$; mutual information estimator $\hat{I}(\cdot;\cdot)$.
\renewcommand{\algorithmicrequire}{ \textbf{HyperParas:}}
\REQUIRE 
learning rate $\beta$; batch size $B$; number of batches $BN$.
\ENSURE 
Trained $f_\omega(\cdot)$.
\FOR{$i=1$ to $BN$}
            \FOR{$\mathcal{G}_j$ in sampled minibatch $\{\mathcal{G}_j\}_{j=1}^{B}$}
            \STATE $(\mathcal{G}_{i,(j,1)},\mathcal{G}_{i,(j,2)})=\operatorname{Algorithm}$ \ref{alg:1} $(\mathcal{G}_{i,j})$;
            \ENDFOR
            \ENDFOR
        \FOR{ep in \#epochs}
        \FOR{$i=1$ to $BN$}
        \STATE $\mathcal{L}=-\frac{1}{B}\sum_{j=1}^B \hat{I}(g_\rho(f_\omega(\mathcal{G}_{i,(j,1)}));g_\rho(f_\omega(\mathcal{G}_{i,(j,2)})))$;
        \STATE $\omega \leftarrow \omega - \beta\nabla_{\omega}\mathcal{L}$; $\rho \leftarrow \rho - \beta\nabla_{\rho}\mathcal{L}$;
        \ENDFOR
        \ENDFOR
\RETURN Trained $f_\omega(\cdot)$;
    \end{algorithmic}
\end{algorithm}
Algorithm \ref{alg:1} describes the details of the automated graph view generator for GIMM. Algorithm \ref{alg:2} and \ref{alg:3} describe the details of the view comparison module for GIMM on node and graph classification, respectively.
Given a graph with $N$ nodes and $M$ edges, where each node has $F$ features.
Asymptotically, the view generator of GIMM requires $O(N^2F+(M+N)F^2)$ floating-point operations (FLOPs), while the view comparison module requires $O(N^2F+NF^2)$ FLOPs.
The FLOPs of AD-GCL, JOAOv2, and AutoGCL are $O(N^2F+(M+N)F^2)$, $O(N^2F+NF^2)$, and $O(N^2F+NF^2)$ FLOPs, respectively.
GIMM's view generator has comparable asymptotic complexity as AD-GCL, and GIMM's view comparison module has comparable asymptotic complexity as JOAO and AutoGCL.
Firstly, the training of the view generator can be completed in a relatively small number of epochs.
Secondly, the training of the view comparison module focuses solely on maximization, while the other three models involve both maximization and minimization.
Thus, GIMM does not require significant computing resources or time.

\bibliographystyle{cas-model2-names}
\bibliography{cas-refs}
\end{document}